\documentclass[11pt, letterpaper]{arxiv}
\usepackage[all]{hypcap}
\usepackage[comma,numbers,sort,compress]{natbib}
\usepackage{hyperref}[citecolor=magenta]
\usepackage[capitalize,noabbrev]{cleveref}
\hypersetup{
    colorlinks = true,
    citecolor = {StanfordRed},
    urlcolor=StanfordRed,
}

\usepackage{microtype}

\usepackage{subcaption}
\usepackage{booktabs} %

\usepackage{algorithm}
\usepackage{algorithmic}
\usepackage{mathrsfs}
\usepackage{setspace}

\usepackage{amsmath}
\usepackage{amssymb}
\usepackage{mathtools}
\usepackage{amsthm}
\definecolor{customColor}{RGB}{255, 204, 102} 

\usepackage{xcolor}

\definecolor{Proprietary}{HTML}{86C2E7}  
\definecolor{OpenSource}{HTML}{96E79F}   

\usepackage[most,skins,theorems]{tcolorbox}
\tcbset{
  aibox/.style={
    width=\linewidth,
    top=7pt,
    bottom=2pt,
    colback=blue!6!white,
    colframe=black,
    colbacktitle=black,
    enhanced,
    center,
    attach boxed title to top left={yshift=-0.1in,xshift=0.15in},
    boxed title style={boxrule=0pt,colframe=white,},
  }
}
\newtcolorbox{AIbox}[2][]{aibox,title=#2,#1}

\newcommand{\scbench}{\textsc{SciVideoBench}\xspace}

\usepackage{wrapfig}
\definecolor{lightblue}{rgb}{0.22,0.45,0.70}%
\usepackage{tabularx}

\usepackage[most]{tcolorbox} 

\definecolor{takeawaybg}{HTML}{F4FCF5} 
\definecolor{takeawayframe}{HTML}{DDF7DF} 

\tcbset{
  takeaway/.style={
    colback=takeawaybg,      
    colframe=takeawayframe,  
    coltitle=black,          
    fonttitle=\bfseries,     
    boxrule=0.8pt,           
    arc=2mm,                 
    left=2mm, right=2mm, top=1mm, bottom=1mm, 
    enhanced,
    sharp corners=all,
    drop shadow              
  }
}

\definecolor{rliableolive}{HTML}{BBCC33}
\definecolor{rliableblue}{HTML}{77AADD}
\definecolor{rliablered}{HTML}{EE8866}

\usepackage{crossreftools}
\usepackage[suppress]{color-edits}

\usepackage{dsfont}
\usepackage{nicefrac}
\newtcolorbox{analysisbox}[1][]{
    enhanced jigsaw,
    colback=white,
    colframe=blue!75!black,
    fonttitle=\bfseries,
    boxsep=5pt,
    left=5pt,
    right=5pt,
    top=5pt,
    bottom=5pt,
    title=#1,
}
\definecolor{editInitialResponse}{RGB}{255, 235, 156} %
\definecolor{editBacktrack}{RGB}{0, 0, 139} %
\definecolor{editRevisedResponse}{RGB}{255, 182, 193} %

\newcommand{\cmark}{\ding{51}}%
\newcommand{\xmark}{\ding{55}}%
\usepackage{pifont}
\usepackage{soul}
\definecolor{highlightmistake}{RGB}{255, 179, 179} 
\definecolor{highlightcorrect}{RGB}{179, 255, 179}

\usepackage[capitalize,noabbrev]{cleveref}

\theoremstyle{plain}

\theoremstyle{definition}

\theoremstyle{remark}

\usepackage[textsize=tiny]{todonotes}

\usepackage{stfloats} 
\usepackage{multirow}
\usepackage{booktabs}
\usepackage{listings}
\usepackage{enumitem}

\usepackage{amsmath,amsfonts,bm}

\def\eqref#1{Eq.~\ref{#1}}

\def\1{\bm{1}}

\DeclareMathAlphabet{\mathsfit}{\encodingdefault}{\sfdefault}{m}{sl}
\SetMathAlphabet{\mathsfit}{bold}{\encodingdefault}{\sfdefault}{bx}{n}

\usepackage{xspace}
\usepackage{xcolor}
\definecolor{visualblue}{RGB}{120, 160, 200}
\definecolor{expertorange}{RGB}{234,  94,  41}
\definecolor{reasoninggreen}{RGB}{44, 102, 47}

\title{\scbench: Benchmarking Scientific Video Reasoning in Large Multimodal Models}

\author{Andong Deng$^{1}$, Taojiannan Yang, Shoubin Yu$^{2}$, Lincoln Spencer$^{1}$,  $~~~~~~~~~~~~~~~~~~~~~~~~~~~~~~~~~~~~~~~~~~~~~~~~~~~~~~~~~~~~~~~~~~~~~~~~~~~~~~$
Mohit Bansal$^{2}$, Chen Chen$^{1}$, Serena Yeung-Levy$^{3}$, Xiaohan Wang$^{3}$ \\
  \vspace{1.0em}
$^{1}$University of Central Florida \\ 
$^{2}$University of North Carolina at Chapel Hill \\
$^{3}$Stanford University \\
    \vspace{1.0em}
\texttt{Link: 
\href{https://scivideobench.github.io/}{Project Page}
~|~
\href{https://github.com/dengandong/SciVideoBench}{Code}
~|~
\href{https://huggingface.co/datasets/groundmore/scivideobench}{HuggingFace}}
}

\begin{document}
\maketitle

\begin{abstract}
\textbf{Abstract:} Large Multimodal Models (LMMs) have achieved remarkable progress across various capabilities; however, complex video reasoning in the scientific domain remains a significant and challenging frontier.
Current video benchmarks predominantly target general scenarios where perception/recognition is heavily relied on, while with relatively simple reasoning tasks, 
leading to saturation and thus failing to effectively evaluate advanced multimodal cognitive skills. To address this critical gap, we introduce \scbench, a rigorous benchmark specifically designed to assess advanced video reasoning in scientific contexts. \scbench consists of 1,000 carefully crafted multiple-choice questions derived from cutting-edge scientific experimental videos spanning over 25 specialized academic subjects and verified by a semi-automatic system. 
Each question demands sophisticated domain-specific knowledge, precise spatiotemporal perception, and intricate logical reasoning, effectively challenging models' higher-order cognitive abilities. 
Our evaluation highlights significant performance deficits in state-of-the-art proprietary and open-source LMMs, including Gemini 2.5 Pro and Qwen2.5-VL, indicating substantial room for advancement in video reasoning capabilities.
Detailed analyses of critical factors such as reasoning complexity and visual grounding provide valuable insights and clear direction for future developments in LMMs, driving the evolution of truly capable multimodal AI co-scientists. We hope \scbench could fit the interests of the community and help to push the boundary of cutting-edge AI for border science. 
\end{abstract}
\abscontent
\begin{figure}[ht]
    \centering
    \includegraphics[width=1.0\textwidth]{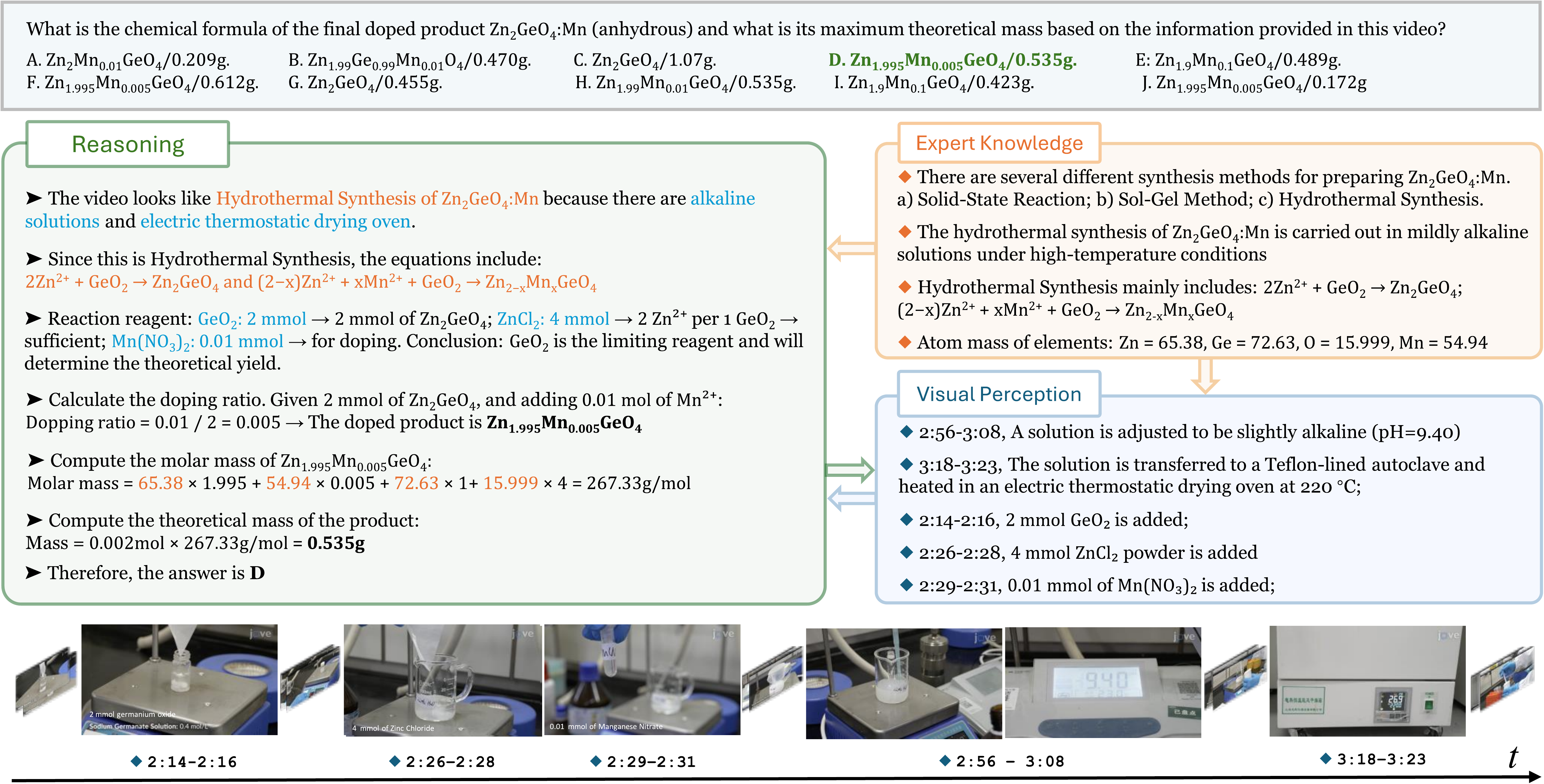}
    
    \caption{\textsc{SciVideoBench} features \textit{research-level} experimental videos accompanied by challenging questions that rigorously evaluate advanced video understanding. It emphasizes the \textit{synergistic interaction} among accurate \textcolor{visualblue}{ visual perception}, \textcolor{expertorange}{expert knowledge}, and sophisticated \textcolor{reasoninggreen}{logical reasoning}.}
    
    \label{fig:pull}
\end{figure}
\vspace{2pt}
\section{Introduction}
\label{sec:intro}

\label{intro}

Large Multimodal Models (LMMs)~\cite{gpt4o2024,anil2023gemini,claude3_2024,lu2024deepseekvl,wang2024qwen2,li2024llavanext} have demonstrated rapid advancements across a diverse range of capabilities, including conversational interaction~\cite{hendrycks2020measuring,yang2018hotpotqa}, code generation~\cite{jimenez2023swebench, sun2024livecodebench}, mathematical reasoning~\cite{cobbe2021training, zhao2025challenging}, and image understanding~\cite{yue2023mmmu,hudson2019gqa}. Among the various input modalities, video presents a particularly rich and complex form of multimodal data, uniquely integrating temporal dynamics, spatial perception, and high-level semantic reasoning. Consequently, video understanding has emerged as a critical frontier, pivotal for advancing LMMs towards next-generation applications in fields such as robotics, interactive education, and scientific discovery.


To evaluate the performance of LMMs, numerous video benchmarks have been developed~\cite{fang2024mmbenchvid,videomme,videommmu,wu2024longvideobench,patraucean2023perception,youcook2,xiao2021next,song2024moviechat,zhao2024mlvu,li2024mvbench,nagrani2024neptune}. However, the majority of existing benchmarks concentrate on relatively general domains—such as movies~\cite{videomme,song2024moviechat}, daily activities~\cite{mangalam2023egoschema,xiao2021next}, and instructional content~\cite{youcook2}—and primarily emphasize tasks like temporal grounding, common-sense reasoning, and event understanding. While these benchmarks once posed significant challenges, current state-of-the-art LMMs, Gemini 2.5 Pro~\cite{google2025gemini25pro}, now exhibit saturated performance, achieving accuracies exceeding 85\% on popular benchmarks like VideoMME~\cite{videomme} and Neptune~\cite{nagrani2024neptune}.

More recent initiatives~\cite{videommmu,song2025videommlu,zhao2025mmvu} have begun to incorporate scientific and educational videos to assess deeper domain knowledge and reasoning. Nevertheless, most of this content remains at a college level, allowing LMMs to achieve success either largely through pure visual recognition with memorized domain knowledge (e.g., asking the virus type given the video) or the visual information does not dominate the reasoning process, thus lacking complex multimodal reasoning. For instance, part of the questions in MMVU~\cite{zhao2025mmvu} leverage visually-irrelevant hypotheses which could enable models to answer without watching the video, e.g., in a video depicting blowing air into a flask of limewater, which reacts with carbon dioxide to form a cloudy calcium carbonate precipitate. The associated question asks:``\emph{Assume that 2.24 liters of gas fully participates in the reaction shown in the video under the standard temperature and pressure conditions, how many grams of precipitate are produced approximately?}''. The assumption about gas volume is actually the most important information to complete the calculation in order to answer the question. Consequently, these benchmarks pose a limited challenge to current proprietary LMMs (Gemini 2.5 Pro achieves 83.6 overall performance on Video-MMMU, and OpenAI o1 achieves 76.1 overall performance on MMVU), and can also be solved via a simple language-based reasoning framework~\cite{zhang2025silvr, yu2025mexa}, which achieves 82.7 and 83.1 on Video-MMMU and Video MMLU, respectively.

Furthermore, current benchmarks inadequately evaluate sophisticated cognitive skills that necessitate expert-level knowledge, intricate logical reasoning, and precise visual perception. Moreover, the videos from these benchmarks are usually from daily life or simple college-level scenarios, while the advanced research-level scenarios, where complex scientific experiments happen, are ignored.

To address this critical gap, we introduce \scbench, an innovative benchmark specifically designed to rigorously assess advanced video reasoning capabilities. \scbench consists of 1,000 meticulously crafted multiple-choice questions derived from research-level experimental videos in physics, chemistry, biology, and medicine that are published with the corresponding journal publications.
These videos are sourced from more than 25 distinct academic subjects, including Fluid Mechanics, Analytical Chemistry, Neuroscience, and Oncology. Each question is categorized into one of three reasoning types: conceptual, hypothetical, or quantitative. As illustrated in Figure.~\ref{fig:pull}, ~\scbench demands that models demonstrate accurate spatio-temporal grounding, possess profound domain knowledge, and execute sophisticated logical reasoning.

We developed \scbench using a multi-stage, agent-human collaborative pipeline. This process involved mining associated experimental manuscripts, leveraging LMMs for initial question generation, and engaging domain experts to validate the question-answer pairs and filter out unanswerable or video-irrelevant questions.  

Our evaluation of both proprietary and open-source LMMs on \scbench reveals consistently low accuracy, underscoring the significant challenge this new benchmark presents and, consequently, the substantial opportunity for advancing research-level
video reasoning capabilities. In-depth analysis of the results reveals that factors such as model architecture, reasoning capacity, and perceptual grounding play a critical role in shaping video reasoning performance.
These insights not only offer clear guidance for future research efforts aimed at developing more sophisticated LMMs but also emphasize the broader potential of \scbench. 
As the first comprehensive research-level video reasoning benchmark, ~\scbench not only provides a rigorous testbed for evaluating current video reasoning abilities, but also serves as a catalyst for innovation, fostering the development of highly capable AI co-scientists that can accelerate future scientific discovery.

\section{Dataset Construction}
\label{data}
\begin{table*}[htb]
    \centering
    \renewcommand{\arraystretch}{0.92}
    \caption{\textbf{Benchmark comparison} for multi-modal video understanding and reasoning tasks.}
    \resizebox{\textwidth}{!}{%
    \begin{tabular}{l| cc | cc | c c }
    \toprule
    \multirow{2}{*}{\textbf{Dataset}} & \multirow{2}{*}{\textbf{Video Domain}} & \multirow{2}{*}{\textbf{Difficulty Level}} & \multirow{2}{*}{\textbf{Knowledge-Driven}} & \multirow{2}{*}{\textbf{Reasoning-Intensive}} & \multirow{2}{*}{\textbf{\#~Ave. Duration (s)}} & \multirow{2}{*}{\textbf{Question Type}} \\
    & & & & & & \\
    \midrule
    MovieChat-1K~\cite{song2024moviechat} & Movie & Elementary & \textcolor{red}{\xmark} & \textcolor{red}{\xmark} & 564 & Open-ended \\
    MLVU~\cite{zhao2024mlvu} & General & Elementary & \textcolor{red}{\xmark} & \textcolor{red}{\xmark} & 930 & Multi-choice \\
    MVBench~\cite{li2024mvbench} & General & Elementary & \textcolor{red}{\xmark} & \textcolor{red}{\xmark} & 16 & Multi-choice \\
    LongVideoBench~\cite{wu2024longvideobench} & General & Elementary & \textcolor{red}{\xmark} & \textcolor{red}{\xmark} & 473 & Multi-choice \\
    TempCompass~\cite{liu2024tempcompass} & General & Elementary & \textcolor{red}{\xmark} & \textcolor{red}{\xmark} & < 30 & Multi-choice \\
    Video-MME~\cite{videomme} & General & Elementary & \textcolor{red}{\xmark} & \textcolor{red}{\xmark} & 1018 & Multi-choice \\
    VSI-Bench~\cite{yang2024thinking} & Embodied & Elementary & \textcolor{red}{\xmark} & \textcolor{red}{\xmark} & 122 & Multi-choice \\
    MMWorld~\cite{he2024mmworld} & General & Elementary & \cmark & \textcolor{red}{\xmark} & 107 & Multi-choice \\
    MMVU~\cite{zhao2025mmvu} & Scientific & College & \cmark & \textcolor{red}{\xmark} & 51 & Multi-choice \\
    Video-MMMU~\cite{videommmu} & Lecture & College & \cmark & \cmark & 506 & Multi-choice \\
    Video-MMLU~\cite{song2025videommlu} & Lecture & College & \cmark & \cmark & 109 & Open-ended \\
    \midrule
    \textbf{SciVideoBench (ours)} & Scientific & Research & \cmark & \cmark & 484 & Multi-choice \\
    \bottomrule
    \end{tabular}%
    }
    \label{tab:benchmark}
    \end{table*}

\subsection{Video Collection and Processing}
To build a high-quality benchmark that rigorously evaluates advanced scientific reasoning in videos, we collect 241 research-grade experimental videos from the \textit{Journal of Visualized Experiments} (JoVE)\footnote{\url{https://www.jove.com}}, a peer-reviewed platform dedicated to publishing methodological videos across a broad spectrum of scientific disciplines. These videos are professionally produced and narratively structured, clearly demonstrating laboratory protocols, scientific phenomena, and technical instrumentation. Their high visual and instructional quality makes them an ideal foundation for constructing a benchmark grounded in authentic scientific practice.

Crucially, each video is accompanied by both a peer-reviewed manuscript and synchronized audio narration. The manuscript describes the experimental protocol and results, while the narration provides temporally aligned explanations of each experimental step as it appears in the video. This tri-modal alignment—\textbf{video}, \textbf{audio}, and \textbf{text}—enables a principled and rigorous approach to question generation and answer verification. The accompanying materials offer detailed procedural descriptions, theoretical context, and experimental outcomes, which are instrumental in formulating meaningful, answerable, and visually grounded questions. This synergy ensures that each question in our benchmark is firmly grounded in both the visual content and its underlying scientific rationale.

We focus our selection on four foundational domains—\textbf{physics}, \textbf{chemistry}, \textbf{biology}, and \textbf{medicine}—which together span a diverse range of procedural complexity and reasoning challenges. Within these domains, videos are carefully selected to ensure the presence of clearly measurable variables (such as reaction time, temperature, or applied force), observable causal relationships and experimental outcomes, as well as logical sequences of actions or instrument usage that facilitate \textit{conceptual}, \textit{quantitative}, and \textit{hypothetical} reasoning.
This targeted curation ensures that each video in \scbench presents rich multimodal cues necessary for rigorous scientific reasoning, making it an ideal testbed for evaluating the capabilities of large multimodal models.

After downloading each video and its corresponding research paper, we use the Whisper~\cite{radford2023robust} automatic speech recognition (ASR) model to extract the audio track and generate a transcript. This transcript is temporally aligned with the visual content to capture step-by-step experimental narration. For the associated paper, we parse and extract the full textual content from the PDF. These multimodal inputs—video, aligned transcript, and paper text—form the basis for downstream question generation. Further details on the preprocessing steps are provided in the Appendix.

\begin{figure}[h]
\centering
\scalebox{1}{\includegraphics[width=1\textwidth]{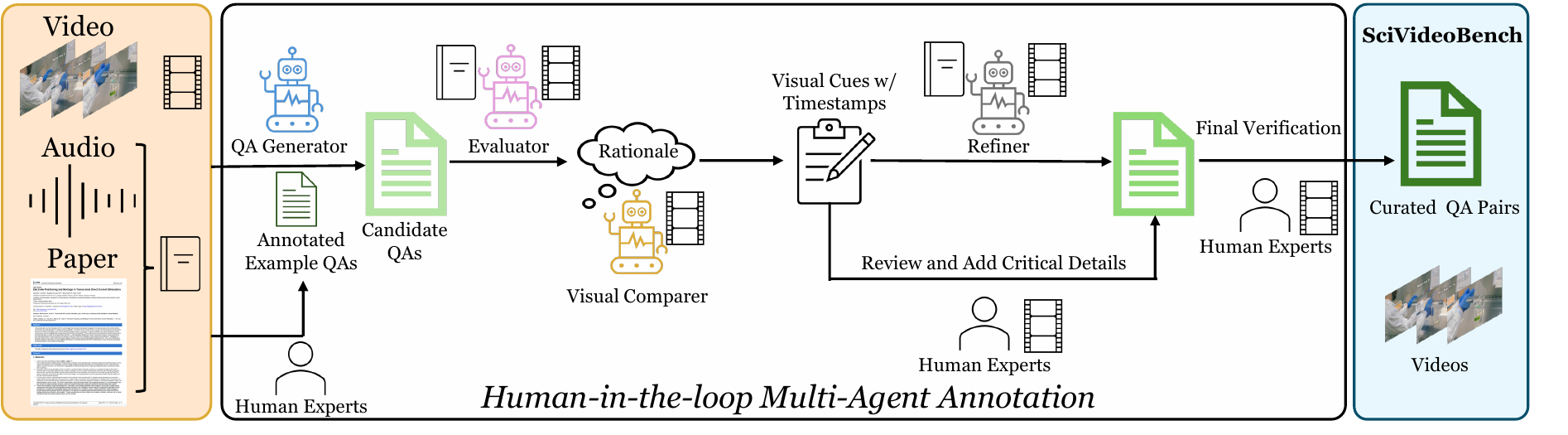}}
\caption{Overview of our annotation pipeline. We manually annotate QA pairs for three example videos using both the video and the associated paper. A multi-agent LLM system generates and refines QA pairs: the QA Generator produces initial questions, the Evaluator answers them with reasoning, the Visual Comparer checks for visual grounding and timestamps cues, and the Refiner ensures questions rely on video content and improves option quality. Human experts verify and refine the final QA pairs. Audio transcripts are omitted for simplicity.}
\label{fig:pipeline}
\end{figure}

\subsection{Annotation Pipeline}

To generate high-quality, research-level question-answer pairs, we develop a semi-automatic annotation pipeline, as shown in Figure~\ref{fig:pipeline}, that integrates video, aligned transcripts, and paired research papers to provide rich scientific and procedural context. This setup enables accurate and visually grounded question generation while reducing manual workload through a multi-agent system.

To ensure quality and guide the semi-automatic annotation process, we first manually annotate a small set of exemplar cases. Four human experts (PhD students from biology, chemistry, medicine, and physics, respectively) review four selected papers (one for each subject) to extract key scientific concepts and procedural steps, then identify corresponding visual cues in the associated videos to construct sophisticated questions. These exemplar QA pairs are further refined and expanded via GPT-4o, resulting in 12 high-quality examples per question type. These are used as prompts to guide the human-in-the-loop multi-agent annotation system.

In the second stage, we assign distinct roles to a set of large language model agents, each responsible for a specific subtask in the annotation pipeline:

\textbf{• QA Generator} produces initial question–answer (QA) pairs from multimodal inputs—video, transcript, and associated paper text—guided by curated exemplar questions. Specifically, we design prompts for Gemini 2.5 Pro to first identify the core scientific question along with key aspects related to theory, experimental operations, and relevant numerical calculations for each video. Using these extracted elements and human-designed exemplars as references, Gemini 2.5 Pro then generates open-ended questions. To expand these into multi-choice format, the model is prompted to create nine scientifically plausible but incorrect distractors for conceptual and hypothetical reasoning, while for quantitative reasoning, distractors are automatically generated by adding appropriate Gaussian noise to the correct answer.

\textbf{• Evaluator} attempts to answer the generated question using the video, the transcript, and paper content, while also producing a detailed rationale that outlines the reasoning process. This step is to ensure all of the questions are answerable given sufficient information.

\textbf{• Visual Comparer} verifies whether the rationale is grounded in the video content by checking that all necessary visual cues are present. It also provides precise timestamps corresponding to each visual reference.

\textbf{• Refiner} refines the QA pairs by replacing generic or descriptive terms in the question with explicit references to visual segments, ensuring that the question cannot be answered without the video content. It also refines the options for each question to ensure the difficulty of the distractors. 

\textbf{• Human Verifier} reviews the refined QA pairs and the outputs from the Visual Comparer to confirm that the question is fully grounded in the visual evidence. Specifically, these verifiers first check the timestamps annotation in questions to make sure the question is answerable based on the video content.  If critical details (e.g., mass or temperature) are mentioned in the transcript but not visible in the video, the expert triggers a script to overlay this information directly onto the relevant video frames. Furthermore, the answer will be scrutinized to ensure that it matches the video content and the reasoning path.

In the final stage, the human experts perform a final review to manually correct any errors and improve clarity. The expert also ensures that each question is closely aligned with the core scientific contribution or focus of the experiment.

\begin{figure}[h]
\centering
\scalebox{1.0}{\includegraphics[width=0.9\textwidth]{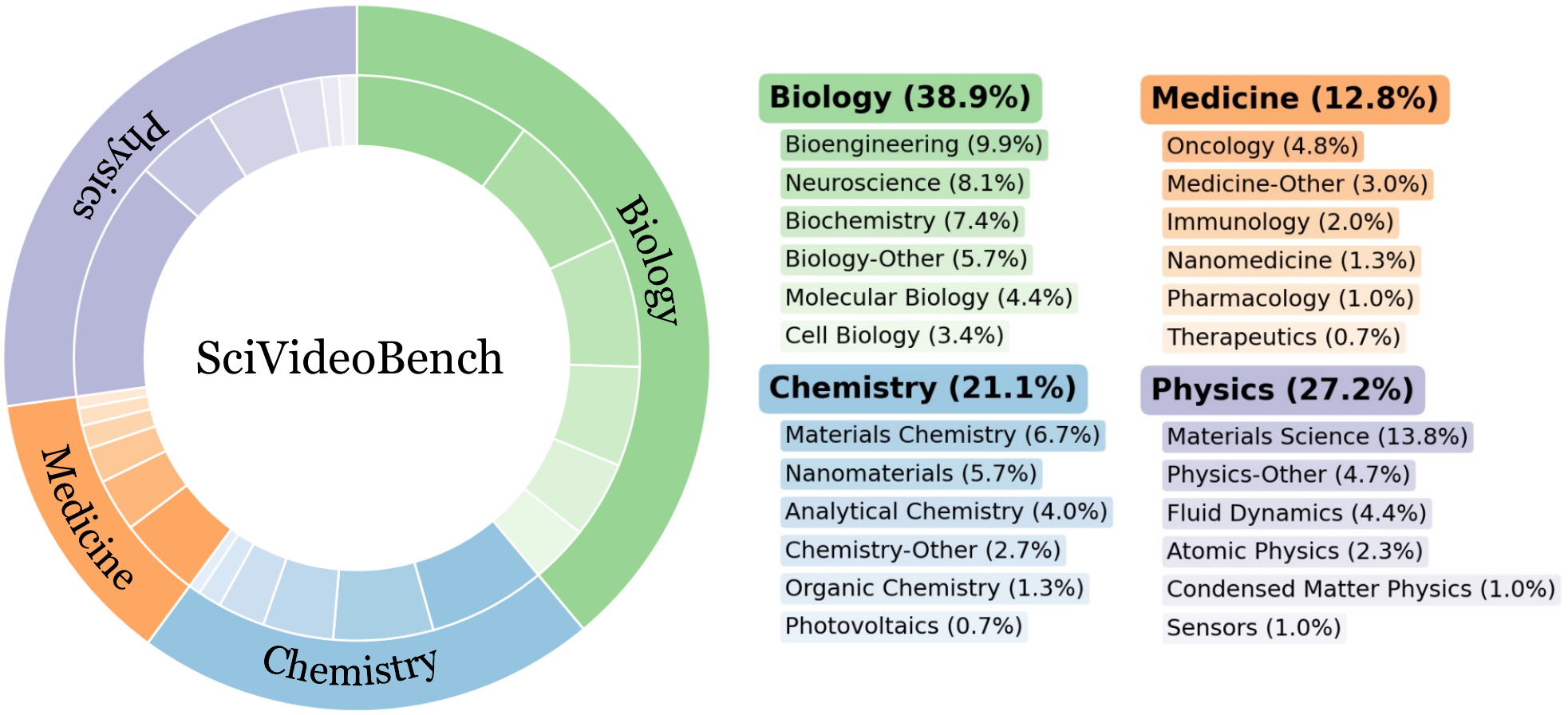}}
\caption{Discipline and subject distribution in ~\scbench. 
Our benchmark covers four major scientific disciplines—Biology, Chemistry, Medicine, and Physics—encompassing more than 25 specialized subjects. 
This diverse coverage ensures a comprehensive evaluation across a wide range of scientific domains.}
\label{fig:datastat}
\end{figure}

\subsection{Statistics}
\begin{figure}[h]
\centering
\begin{subfigure}{0.32\textwidth}
    \includegraphics[width=\linewidth]{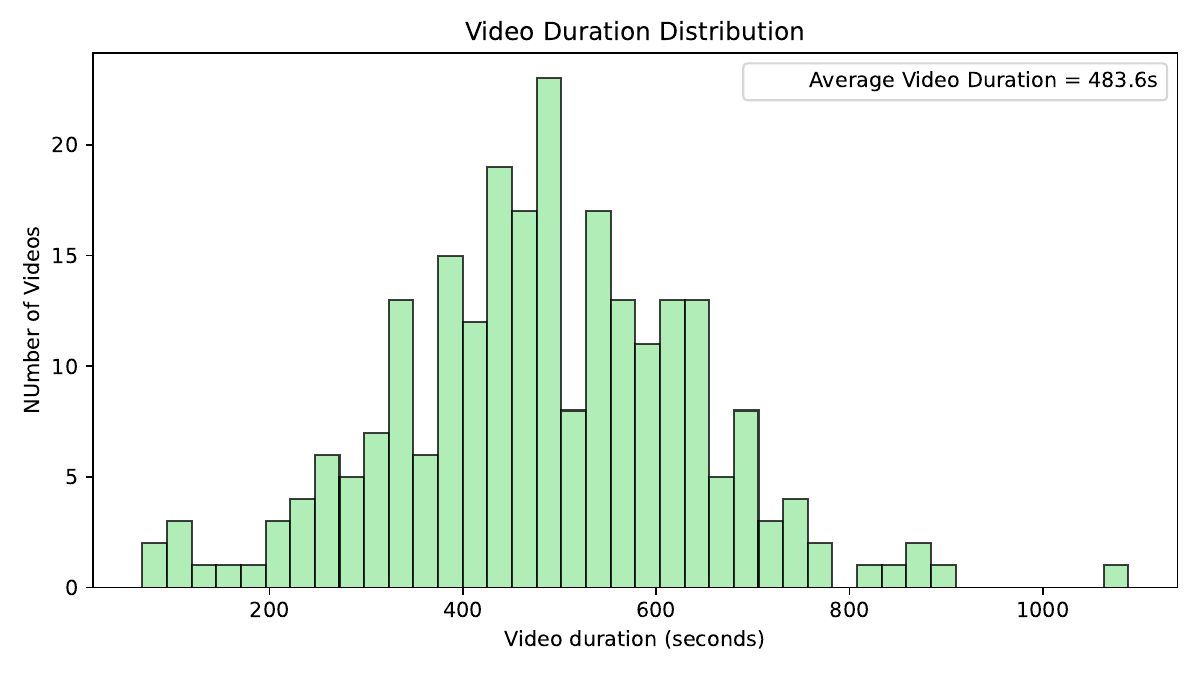}
    \caption{Video duration distribution.}
    \label{fig:video_duration}
\end{subfigure}\hfill
\begin{subfigure}{0.32\textwidth}
    \includegraphics[width=\linewidth]{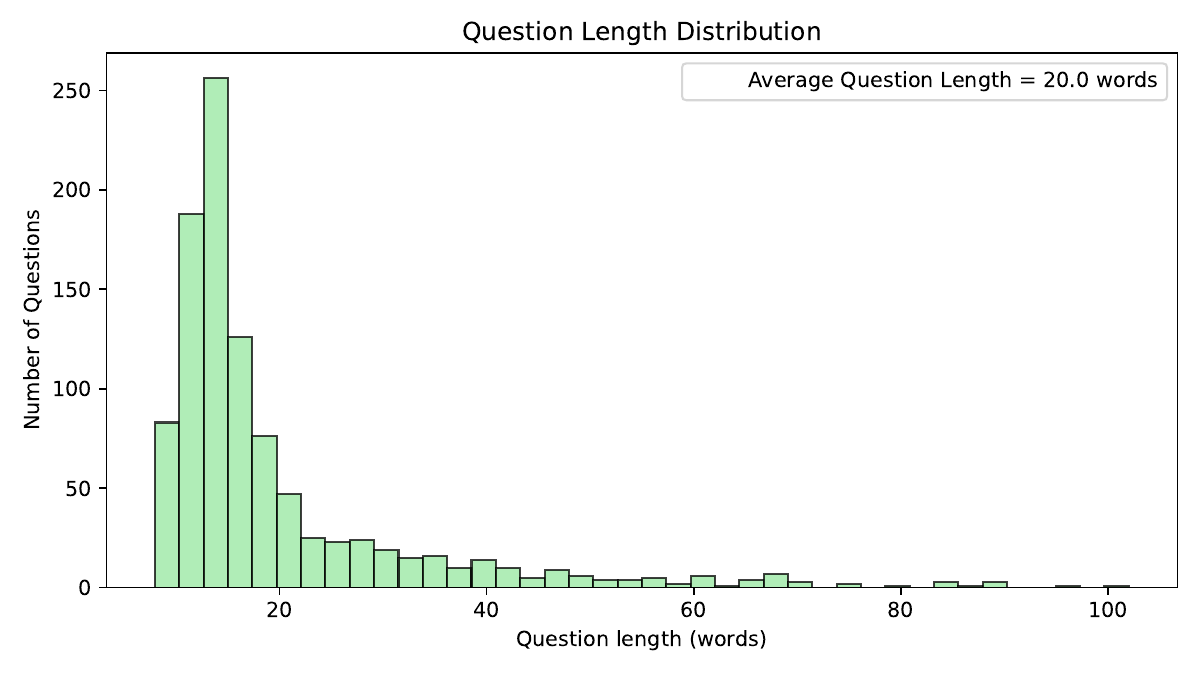}
    \caption{Question length distribution.}
    \label{fig:question_length}
\end{subfigure}\hfill
\begin{subfigure}{0.32\textwidth}
    \includegraphics[width=\linewidth]{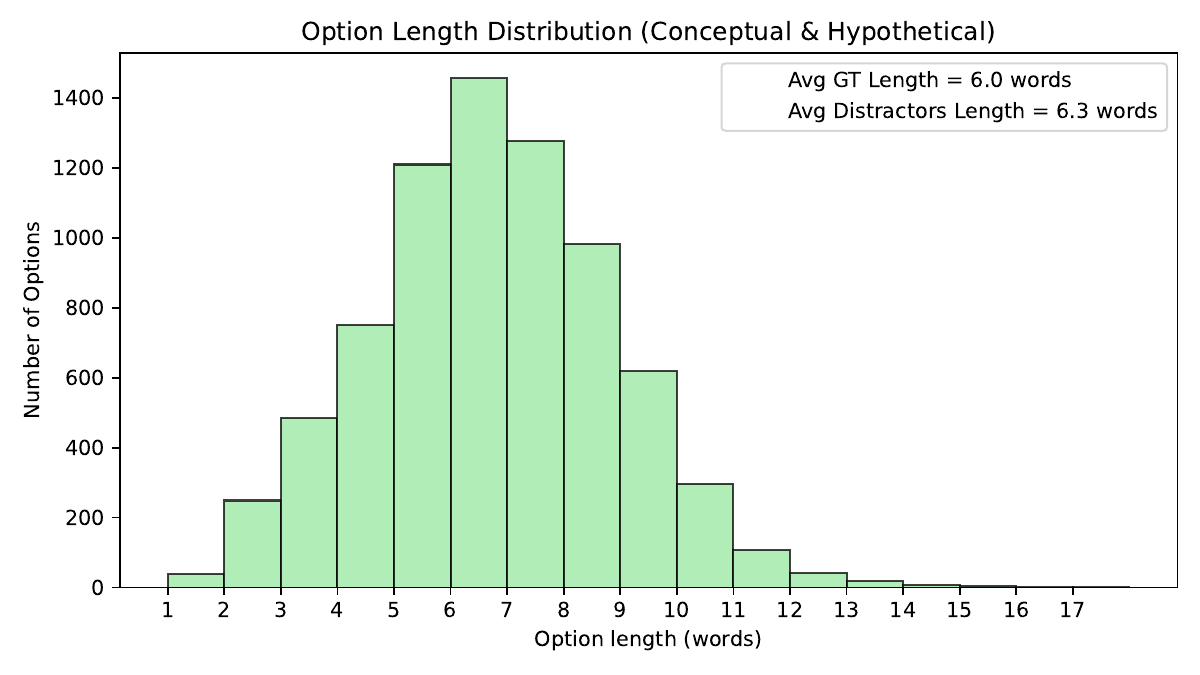}
    \caption{Option length distribution.}
    \label{fig:option_length}
\end{subfigure}
\caption{Dataset statistics in ~\scbench: (a) video duration, (b) question length, and (c) option length distributions. 
These statistics provide a comprehensive overview of the dataset’s temporal scale and linguistic diversity.}
\label{fig:bench_stat}
\end{figure}

We collected a total of 241 experimental videos spanning four major domains: Physics, Chemistry, Biology, and Medicine. These videos cover more than 25 distinct scientific subjects, as illustrated in Figure~\ref{fig:datastat}. The videos in ~\scbench have a competitive average duration of 484 seconds, with the duration distribution shown in Figure~\ref{fig:bench_stat}(a). This relatively long temporal scale ensures that the benchmark reflects the complexity and extended reasoning often required in real-world scientific experiments.

\begin{figure}[H]
\centering
\vspace{-0.2cm}
\scalebox{1.0}{\includegraphics[width=\textwidth]{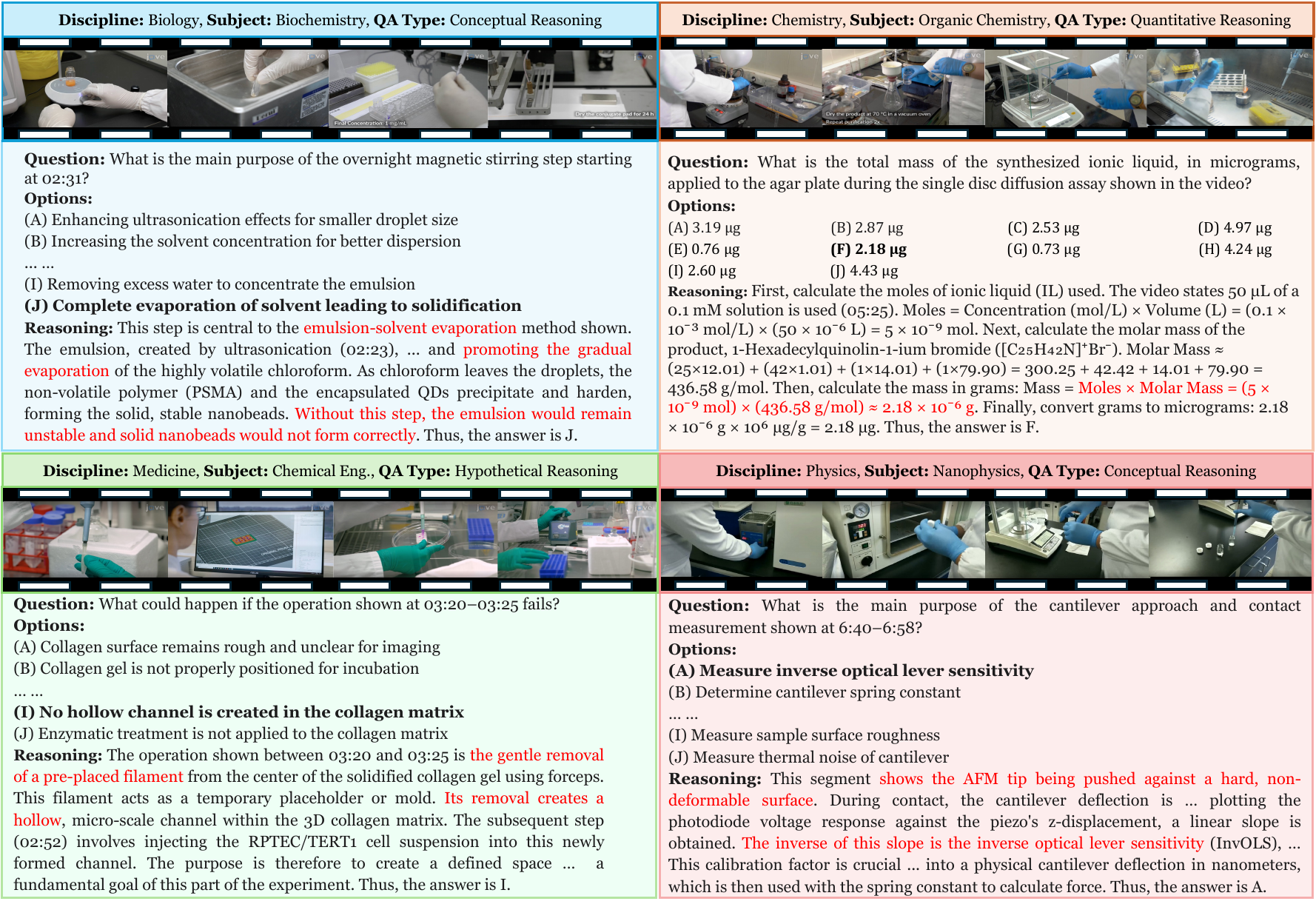}}
\caption{Examples of \scbench, including videos from 4 disciplines (Physics, Biology, Chemistry, and Medicine), which involve 19 different subjects. The research-level QAs challenge LMMs in three different aspects (\textit{Conceptual, Hypothetical, and Quantitative}) that are of vital importance in scientific experiment video understanding.}
\label{fig:example}
\vspace{-10pt}
\end{figure}

Building upon these videos, we annotated a total of 1,000 challenging questions that demand research-level knowledge for both perception and reasoning. As shown in Figure~\ref{fig:bench_stat}(b), the average question length demonstrates that the questions are more linguistically complex than those in existing benchmarks, reinforcing the emphasis on detailed experimental understanding. Furthermore, the option length distribution in Figure~\ref{fig:bench_stat}(c) highlights the balanced design of ground-truth answers and distractors, with comparable average lengths, thereby reducing annotation bias and ensuring fair evaluation. To further capture the nature of academic research and experimental analysis, we carefully designed three distinct question types that reflect common reasoning scenarios observed across the videos, as illustrated in Figure~\ref{fig:example}. Collectively, these design choices ensure that the question–answer pairs in ~\scbench require deeper domain expertise and multi-step reasoning, distinguishing it from prior benchmarks that mainly target elementary or undergraduate-level understanding.


\begin{itemize}[leftmargin=0.15in]
    \setlength\itemsep{-0.25em}
    \item \textbf{245 Quantitative Reasoning} involves numeric perception, reasoning, and calculation. For a specific quantitative question, we require that all the numeric information must come from the video, which naturally ask the model to perceive informative values from the video before perform complicated calculations and reasoning.
    \item \textbf{385 Hypothetical Reasoning} focuses on specific experimental operations that play a critical role for the motivation or outcome of the whole experiment. This question type usually involves hypothesized error, what-if analysis, or experimental control logic, which requires both expert-level knowledge on a specific domain and accurate visual perception to capture important details in the videos.
    \item \textbf{370 Conceptual Reasoning} explore the mechanisms, protocols, and scientific principles behind the operations in the experiment that are inferred via visual/operational cues. 
\end{itemize}

\section{Experiments}
\label{exp}
\subsection{Models}
\noindent\textbf{Vision-Blind Baselines} To investigate the importance of visual information of our benchmark, we first conduct the vision-blind evaluation using GPT-4o~\cite{gpt4o2024} and Qwen2.5 series~\cite{qwen2.5}.

\noindent\textbf{Proprietary Models} We evaluate five proprietary LLMs that have superior performance over other popular video benchmarks, including Gemini-1.5-Pro~\cite{anil2023gemini}, Gemini-2.0-Flash~\cite{anil2023gemini}, Gemini-2.5-Pro~\cite{google2025gemini25pro}, and GPT-4o~\cite{gpt4o2024}. 

\noindent\textbf{Open-Source Models} We comprehensively evaluate 30 open-sourced LMMs with the parameter volume ranging from as tiny as 0.5B to 78B, including Qwen-VL series~\cite{bai2025qwen2.5vl,wang2024qwen2}, InternVL series~\cite{internvl2,zhu2025internvl3,internvl2.5}, InternVide2.5-Chat-8B~\cite{internvideo2.5}, LLaVA-OneVision Series~\cite{li2024llavaov}, LLaVA-NeXT-Video-32B series~\cite{li2024llavanext} and LongVA~\cite{zhang2024longva}.


\subsection{Evaluation Setting}
We adopt the default frame sampling strategy for all open-source models. Specifically, we sample 768 frames for Qwen-2.5-VL~\cite{bai2025qwen2.5vl}, 512 frames for InternVideo2.5-Chat-8B~\cite{internvideo2.5}, 128 frames for LongVA~\cite{zhang2024longva}, 16 frames for InternVL2~\cite{internvl2}, and 32 frames for InternVL3~\cite{zhu2025internvl3}, LLaVA-OneVision~\cite{li2024llavaov}, and LLaVA-NeXT-Video-32B~\cite{li2024llavanext}. For Gemini~\cite{anil2023gemini,comanici2025gemini,google2025gemini25pro}, we set the frame rate to 1 FPS, and for GPT-4o~\cite{gpt4o2024} we use 256 frames. The temperature is fixed to 0 to ensure stable and reproducible evaluation results. All evaluations are using the LMM-Eval toolkit~\cite{zhang2024lmmsevalrealitycheckevaluation} and are conducted on 8 NVIDIA H100 GPUs.

\subsection{Human Evaluation}
We also conduct a human evaluation by recruiting three graduate students from each discipline to answer questions in a closed-book setting. Their overall accuracy is only 17.4\% (Table~\ref{tab:main}), specifically, the quantitative reasoning is only 14.29\%, showing that even advanced students cannot handle the benchmark, which requires research-level expertise.

\begin{table*}[htbp]
\centering
\caption{Evaluation Results of Proprietary Models and Open-Source Models on \scbench.}
\resizebox{\textwidth}{!}{%
\begin{tabular}{@{}llllccccccc@{}}

\toprule
\multirow{2}{*}{\textbf{Models}} & \multirow{2}{*}{\textbf{LLM}} & \multirow{2}{*}{\textbf{Size}} & 
\multirow{2}{*}{\textbf{Overall}} & 
\multicolumn{3}{c}{\textbf{Question Type}} & 
\multicolumn{4}{c}{\textbf{Discipline}} \\
\cmidrule(lr){5-7} \cmidrule(lr){8-11}
& & & & \textbf{Conceptual} & \textbf{Hypothetical} & \textbf{Quantitative} & 
\textbf{Biology} & \textbf{Chemistry} & \textbf{Medicine} & \textbf{Physics} \\
\midrule
\rowcolor{gray!20} \multicolumn{11}{c}{{\textit{Random Guess}}}   \\
- & - & - & 10.00 & 10.00 & 10.00 & 10.00 & 10.00 & 10.00 & 10.00 & 10.00 \\
\midrule
\rowcolor{gray!20} \multicolumn{11}{c}{{\textit{Human Evaluation (Graduate Students)}}}   \\
- & - & - & 17.40 & 18.11 & 18.70 & 14.29 & 15.88 & 16.06 & 21.19 & 18.88 \\
\midrule
\rowcolor{gray!20} \multicolumn{11}{c}{{\textit{Vision-Blind Baselines} }}   \\
GPT-4o~\cite{gpt4o2024} & - & - & 15.80 & 14.05 & 20.00 & 11.84 & 17.53 & 13.95 & 16.13 & 14.33 \\ \midrule
 & - & 0.5B & 12.40 & 11.35 & 13.51 & 12.24 & 13.20 & 13.94 & 11.21 & 10.97 \\
 & - & 1.5B & 13.40 & 11.62 & 19.22 & 6.94 & 14.43 & 12.73 & 14.95 & 11.91 \\
 & - & 3B & 16.40 & 17.03 & 20.52 & 8.98 & 15.89 & 16.36 & 15.89 & 17.24 \\
Qwen2.5~\cite{qwen2.5} & - & 7B & 16.70 & 18.65 & 19.74 & 8.98 & 18.34 & 10.91 & 14.95 & 18.18 \\
 & - & 32B & 17.10 & 18.11 & 21.30 & 8.98 & 19.32 & 16.97 & 13.08 & 15.67 \\
 & - & 72B & 18.90 & 21.89 & 18.44 & 15.10 & 19.07 & 18.79 & 20.56 & 18.18 \\

\midrule
\rowcolor{gray!20} \multicolumn{11}{c}{{\textit{Proprietary Models}}}   \\
Gemini-2.5-Pro~\cite{google2025gemini25pro} & - & - & \textbf{64.30} & \textbf{69.73} & \textbf{67.79} & 50.61 & \textbf{64.79} & \textbf{61.82} & \textbf{74.77} & \textbf{61.44} \\
Gemini-2.5-Flash~\cite{comanici2025gemini} & - & - & 46.40 & 50.81 & 44.16 & 43.27 & 44.01 & 49.70 & 55.14 & 44.83 \\
Gemini-1.5-Pro~\cite{anil2023gemini} & - & - & 27.50 & 27.84 & 28.31 & 25.71 & 27.38 & 26.06 & 27.10 & 28.53 \\
Gemini-2.0-Flash~\cite{anil2023gemini} & - & - & 25.70 & 28.38 & 24.94 & 22.86 & 24.69 & 26.06 & 22.43 & 27.90 \\
GPT-4o~\cite{gpt4o2024} & - & - & 24.90 & 30.27 & 28.05 & 11.84 & 21.52 & 29.70 & 31.78 & 24.45 \\
\midrule
\rowcolor{gray!20} \multicolumn{11}{c}{{\textit{Proprietary Models w/ Chain-of-Thought Reasoning }}}   \\
Gemini-1.5-Pro~\cite{anil2023gemini} & - & - & 48.60\textsubscript{\textcolor{green!50!black}{(+21.10)}} & 47.03\textsubscript{\textcolor{green!50!black}{(+19.19)}} & 48.57\textsubscript{\textcolor{green!50!black}{(+20.26)}} & \textbf{51.02}\textsubscript{\textcolor{green!50!black}{(+25.31)}} & 52.60\textsubscript{\textcolor{green!50!black}{(+25.22)}} & 44.19\textsubscript{\textcolor{green!50!black}{(+18.13)}} & 54.84\textsubscript{\textcolor{green!50!black}{(+27.74)}} & 44.78\textsubscript{\textcolor{green!50!black}{(+16.25)}} \\
Gemini-2.0-Flash~\cite{anil2023gemini} & - & - & 39.70\textsubscript{\textcolor{green!50!black}{(+14.00)}} & 39.19\textsubscript{\textcolor{green!50!black}{(+10.81)}} & 39.74\textsubscript{\textcolor{green!50!black}{(+14.80)}} & 40.41\textsubscript{\textcolor{green!50!black}{(+17.55)}} & 41.56\textsubscript{\textcolor{green!50!black}{(+16.87)}} & 40.70\textsubscript{\textcolor{green!50!black}{(+14.64)}} & 35.48\textsubscript{\textcolor{green!50!black}{(+13.05)}} & 37.01\textsubscript{\textcolor{green!50!black}{(+9.11)}} \\
GPT-4o~\cite{gpt4o2024} & - & - & 35.00\textsubscript{\textcolor{green!50!black}{(+10.10)}} & 37.84\textsubscript{\textcolor{green!50!black}{(+7.57)}} & 32.73\textsubscript{\textcolor{green!50!black}{(+4.68)}} & 34.29\textsubscript{\textcolor{green!50!black}{(+22.45)}} & 31.78\textsubscript{\textcolor{green!50!black}{(+10.26)}} & 37.58\textsubscript{\textcolor{green!50!black}{(+7.88)}} & 36.45\textsubscript{\textcolor{green!50!black}{(+4.67)}} & 37.30\textsubscript{\textcolor{green!50!black}{(+12.85)}} \\

\midrule
\rowcolor{gray!20} \multicolumn{11}{c}{{\textit{Open-Source Models w/ Chain-of-Thought Reasoning}}}   \\
InternVL-3-78B~\cite{zhu2025internvl3} & Qwen2.5 & 72B & 37.90 & 51.62 & 35.58 & 20.82 & 38.39 & 36.36 & 38.32 & 37.93 \\
InternVL-3-14B~\cite{zhu2025internvl3} & Qwen2.5 & 14B & {34.20} & 46.49 & 30.65 & 21.22 & 35.70 & 30.30 & 37.38 & 33.23 \\
InternVL-3-14B-Instruct~\cite{zhu2025internvl3} & Qwen2.5 & 14B & 31.50 & 42.43 & 27.79 & 20.82 & 30.07 & 32.73 & 37.38 & 30.72 \\
InternVL-3-8B~\cite{zhu2025internvl3} & Qwen2.5 & 7B & 25.50 & 34.32 & 24.16 & 14.29 & 25.18 & 24.85 & 28.97 & 25.08 \\
InternVL2-Llama3-76B~\cite{internvl2} & Hermes2 & 70B & 24.90 & 27.30 & 22.34 & 25.31 & 25.18 & 24.24 & 28.97 & 23.51\\
InternVL-3-9B-Instruct~\cite{zhu2025internvl3} & InternLM3 & 8B & 24.00 & 29.19 & 22.08 & 19.18 & 25.92 & 27.88 & 17.76 & 21.63 \\
InternVL-3-2B~\cite{zhu2025internvl3} & Qwen2.5 & 1.5B & 22.20 & 30.00 & 21.82 & 11.02 & 21.76 & 16.36 & 25.23 & 24.76 \\
Qwen2.5-VL-32B-Instruct~\cite{bai2025qwen2.5vl} & Qwen2.5 & 32B & 20.80 & 20.54 & 22.86 & 17.96 & 20.78 & 18.18 & 16.82 & 23.51 \\
LLaVA-OneVision-7B~\cite{li2024llavaov} & Qwen2.5 & 7B & 19.90 & 24.32 & 20.52 & 12.24 &16.87 & 21.21 & 28.04 & 20.38 \\
Qwen2.5-VL-3B-Instruct~\cite{bai2025qwen2.5vl} & Qwen2.5 & 3B & 18.10 & 19.19 & 18.96 & 15.10 & 17.60 & 18.79 & 19.63 & 17.87 \\
InternVL-3-1B~\cite{zhu2025internvl3} & Qwen2.5 & 0.5B & 14.00 & 16.76 & 13.51 & 10.61 & 15.40 & 13.94 & 14.02 & 12.23 \\
LLaVA-OneVision-0.5B~\cite{li2024llavaov} & Qwen2.5 & 0.5B & 10.40 & 11.08 & 10.65 & 8.98 & 8.56 & 12.12 & 13.08 & 10.97 \\
\midrule
\rowcolor{gray!20} \multicolumn{11}{c}{{\textit{Open-Source Models} (0.5B - 4B)}}   \\
InternVL-3-2B-Instruct~\cite{zhu2025internvl3} & Qwen2.5 & 1.5B & 24.00 & 31.08 & 23.90 & 13.47 & 24.21 & 21.82 & 23.36 & 25.08 \\
InternVL-3-2B~\cite{zhu2025internvl3} & Qwen2.5 & 1.5B & 22.90 & 31.08 & 22.60 & 11.02 & 21.52 & 21.21& 24.30 & 25.08 \\
InternVL2-4B~\cite{internvl2} & Phi-3-mini & 4B & 21.30 & 26.22 & 24.16 & 9.39 & 18.09 & 24.85 & 30.84 & 20.38 \\
InternVL-3-1B-Instruct~\cite{zhu2025internvl3} & Qwen2.5 & 0.5B & 18.90 & 26.76 & 18.44 & 7.76 & 18.34 & 18.18 & 22.43 & 18.81 \\
InternVL-3-1B~\cite{zhu2025internvl3} & Qwen2.5 & 0.5B & 18.50 & 25.95 & 18.44 & 7.35 & 17.60 & 16.36 & 25.23 & 18.50 \\
Qwen2.5-VL-3B-Instruct~\cite{bai2025qwen2.5vl} & Qwen2.5 & 3B & 16.10 & 20.00 & 17.92 & 7.35 & 15.16 & 13.94 & 18.69 & 17.55 \\
InternVL2-1B~\cite{internvl2} & InternLM2 & 0.5B & 14.40 & 17.57 & 14.03 & 10.20 & 12.96 & 13.94 & 16.82 & 15.67 \\
InternVL2-2B~\cite{internvl2} & InternLM2 & 2B & 13.10 & 14.59 & 15.06 & 7.76 & 11.98 & 12.12 & 12.15 & 15.36 \\
LLaVA-OneVision-0.5B~\cite{li2024llavaov} & Qwen2.5 & 0.5B & 12.10 & 15.41 & 12.99 & 5.71 & 11.74 & 13.94 & 13.08 & 11.29 \\

\rowcolor{gray!20} \multicolumn{11}{c}{{\textit{Open-Source Models} (7B - 14B)}}   \\
InternVL-3-14B~\cite{zhu2025internvl3} & Qwen2.5 & 14B & 35.70 & 53.51 & 35.32 & 9.39 & 35.94 & 33.94 & 38.32 & 35.42 \\
InternVL-3-14B-Instruct~\cite{zhu2025internvl3} & Qwen2.5 & 14B & 35.70 & 53.24 & 36.36 & 8.16 & 35.70 & 34.55 & 38.32 & 35.42 \\
InternVL-3-8B~\cite{zhu2025internvl3} & Qwen2.5 & 7B & 30.50 & 44.59 & 30.39 & 9.39 & 29.10 & 31.52 & 35.51 & 30.09 \\
InternVL-3-8B-Instruct~\cite{zhu2025internvl3} & Qwen2.5 & 7B & 29.40 & 43.78 & 29.35 & 7.76 & 26.16 & 31.52 & 34.58 & 30.72 \\
InternVL-3-9B-Instruct~\cite{zhu2025internvl3} & InternLM3 & 8B & 29.20 & 40.27 & 30.91 & 9.80 & 29.10 & 29.70 & 33.64 & 27.69 \\
InternVL-3-9B~\cite{zhu2025internvl3} & InternLM3 & 8B & 27.20 & 38.65 & 27.79 & 8.98 & 25.67 & 26.06 & 31.78 & 28.21 \\
InternVideo2.5-Chat-8B~\cite{internvideo2.5} & Qwen2.5 & 7B & 25.30 & 37.84 & 23.12 & 9.80 & 20.29 & 23.64 & 31.78 & 30.41 \\
InternVL2-8B~\cite{internvl2} & InternLM2 & 7B & 19.40 & 24.86 & 18.96 & 11.84 & 17.85 & 21.21 & 19.63 & 20.38 \\
LLaVA-OneVision-7B~\cite{li2024llavaov} & Qwen2.5 & 7B & 18.80 & 23.51 & 19.22 & 11.02 & 15.56 & 23.03 & 26.17 & 18.18 \\
Qwen2.5-VL-7B-Instruct~\cite{bai2025qwen2.5vl} & Qwen2.5 & 7B & 16.40 & 18.92 & 17.14 & 11.43 & 16.87 & 15.15 & 16.82 & 16.30 \\
LongVA~\cite{zhang2024longva} & Qwen2 & 7B & 14.30 & 16.15 & 14.94 & 10.31 & 13.69 & 15.00 & 15.65 & 14.11 \\

\rowcolor{gray!20} \multicolumn{11}{c}{{\textit{Open-Source Models} (26B - 40B)}}   \\
InternVL-3-38B~\cite{zhu2025internvl3} & Qwen2.5 & 32B & 38.30 & 53.78 & 38.44 & 14.69 & 36.67 & 40.00 & 42.06 & 38.24 \\
InternVL-3-38B-Instruct~\cite{zhu2025internvl3} & Qwen2.5 & 32B & 37.30 & 52.43 & 37.14 & 14.69 & 35.94 & 39.39 & 40.19 & 36.99  \\
InternVL2-40B~\cite{internvl2} & Hermes2 & 34B & 23.80 & 28.38 & 23.64 & 17.14 & 22.74 & 21.82 & 30.84 & 23.82\\
Qwen2.5-VL-32B-Instruct~\cite{bai2025qwen2.5vl} & Qwen2.5 & 32B & 21.50 & 24.86 & 24.68 & 11.43 & 22.49 & 16.36 & 20.56 & 23.20 \\
LLaVA-NeXT-Video-32B~\cite{li2024llavanext} & Qwen2 & 32B & 21.10 & 26.22 & 22.86 & 10.61 & 19.80 & 22.42 & 23.36 & 21.32 \\
InternVL2-26B~\cite{internvl2} & InternLM2 & 20B & 19.50 & 21.89 & 20.26 & 14.69 & 18.83 & 18.18 & 22.43 & 20.06 \\

\rowcolor{gray!20} \multicolumn{11}{c}{{\textit{Open-Source Models} (> 70B)}}   \\
InternVL-3-78B-Instruct~\cite{zhu2025internvl3} & Qwen2.5 & 72B & 38.80 & 57.30 & 39.74 & 9.39 & 37.90 & 39.39 & 46.73 & 36.99 \\
InternVL-3-78B~\cite{zhu2025internvl3} & Qwen2.5 & 72B & 38.50 & 56.76 & 39.22 & 9.80 & 37.65 & 37.58 & 46.73 & 37.30 \\
InternVL2-Llama3-76B~\cite{internvl2} & Hermes2 & 70B & 26.30 & 28.38 & 27.01 & 22.04 & 24.94 & 29.70 & 29.91 & 25.08\\
Qwen2.5-VL-72B-Instruct~\cite{bai2025qwen2.5vl} & Qwen2.5 & 72B & 20.30 & 21.62 & 20.78 & 17.55 & 21.27 & 16.97 & 24.30 & 19.44 \\

\bottomrule
\label{tab:main}
\end{tabular}
}
\end{table*}

\subsection{Blind Baseline Results} To assess the role of visual information in \scbench, we evaluate blind baselines by removing the video input and feeding only textual content to the models. As shown in Table~\ref{tab:main}, GPT-4o achieves only 15.80\% overall accuracy, with particularly poor results on Quantitative Reasoning (11.84\%), barely above random guess. Qwen2.5 exhibits similar trends across scales: small variants (0.5B–3B) remain near chance, while larger models (32B and 72B) reach at most 18.90\%. Despite moderate gains from scaling, both GPT-4o and Qwen2.5 fail to exceed 20\% accuracy without visual input. Together with the time-specific, observation-heavy expressions, e.g., "starting at 02:31", in the questions shown in Figure~\ref{fig:example}, these results underscore that \textbf{visual information is indispensable for solving \scbench}, especially for reasoning types requiring precise observation and measurement from the video content.

\subsection{Proprietary Models vs. Open-Source Models}
As shown in Table~\ref{tab:main}, proprietary models substantially outperform open-source counterparts on our \scbench. The strongest proprietary system, Gemini-2.5-Pro, achieves 64.30\% overall accuracy, whereas the best open-source model, InternVL-3-78B-Instruct, reaches only 38.80\%. The gap is especially pronounced in Quantitative Reasoning, where Gemini-2.5-Pro attains 50.61\%—more than double the best open-source score (22.04\% by InternVL2-Llama3-76B). Nevertheless, top open-source models demonstrate competitiveness against earlier proprietary models: for example, most InternVL-3 variants outperform Gemini-1.5-Pro, and even InternVL-3-2B surpasses it on Conceptual Reasoning.
At the lower end, performance drops sharply. The weakest open-source model, LLaVA-OneVision-0.5B, achieves just 12.10\% overall, with its Quantitative score (5.71\%) falling below even the blind GPT-4o baseline (11.84\%). On average, \textbf{proprietary models achieve nearly double the accuracy of open-source models (38.83\% vs. 18.14\%)}, highlighting the considerable advantage of proprietary systems in handling the multimodal and reasoning-intensive demands of \scbench.

\begin{tcolorbox}[takeaway,title=Takeaway: Proprietary vs. Open-Source Models]
Gemini-2.5-Pro leads overall. Proprietary models dominate quantitative reasoning, while top open-source models remain competitive on conceptual/hypothetical tasks.
\end{tcolorbox}

\subsection{Results of Different Question Types}
As shown in the right portion of Table~\ref{tab:main}, Quantitative Reasoning consistently emerges as the most challenging category across models, regardless of architecture or scale. This difficulty is evident not only in proprietary models but also in open-source models, where Quantitative scores are systematically lower than those for Conceptual or Hypothetical reasoning. In fact, several open-source models perform close to or even below the random-guess baseline, for example, InternVL2-4B achieves only 9.39\%, and InternVL-3-1B-Instruct reaches 7.76\% in Quantitative Reasoning. By contrast, Conceptual Reasoning generally yields the highest scores, slightly ahead of Hypothetical Reasoning, although the gap is modest. These patterns indicate that \scbench places particularly high demands on precise numerical reasoning and multi-step calculation abilities, while Conceptual and Hypothetical questions—though still challenging—tend to be more approachable for current multimodal LLMs.

\subsection{Results of Different Disciplines}
Across different disciplines, the performance trends remain relatively consistent, indicating that no single discipline presents a disproportionately higher level of difficulty in \scbench. For Gemini-2.5-Pro, Chemistry emerges as the lowest-performing discipline, whereas Medicine achieves the highest accuracy; while Gemini-1.5-Pro displays the opposite pattern, with its best results in Chemistry and comparatively lower performance in other disciplines. This contrast suggests that the impact of model architecture and scale can vary across domains, and that strengths in one discipline do not necessarily translate directly to others.

\subsection{The Impact of Chain-of-Thought Prompting}
\begin{figure}[H]
\centering
\vspace{-0.2cm}
\scalebox{1.0}{\includegraphics[width=\textwidth]{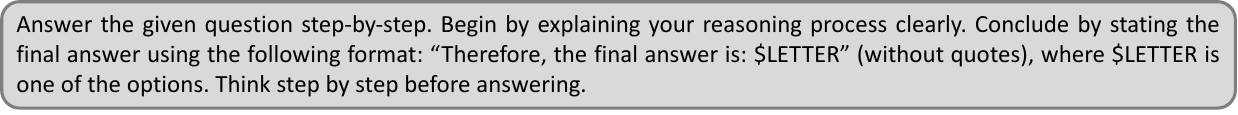}}
\caption{Chain-of-thought prompt used in \scbench}
\label{fig:cot_prompt}
\vspace{-10pt}
\end{figure}

To highlight the reasoning-intensive nature of our \scbench benchmark and to examine the performance gains attributable to deeper reasoning chains, we further evaluate models using chain-of-thought (CoT) prompts as shown in Figure~\ref{fig:cot_prompt}, following the methodology in~\cite{zhao2025mmvu}. As shown in Table~\ref{tab:main}, models with CoT prompting consistently and substantially outperform their vanilla counterparts. 
Gemini-1.5-Pro shows the most pronounced improvement, with overall accuracy rising by \textbf{+21.10\%}, and Quantitative Reasoning achieving a remarkable \textbf{+25.31\%} boost—surpassing even the stronger Gemini-2.5-Pro. Gemini-2.0-Flash and GPT-4o also exhibit substantial overall gains (+14.00\% and +10.10\%, respectively). Notably, GPT-4o’s Quantitative Reasoning accuracy increases from a near-random 11.84\% to 34.29\%, underscoring the transformative impact of explicit reasoning steps.
Across reasoning categories, the magnitude of improvement is not uniform. On average, Quantitative Reasoning benefits the most from the thinking mode (+21.77\%), while Conceptual Reasoning shows the smallest relative gain (+12.52\%). This disparity suggests that the benefits of CoT are particularly strong in scenarios requiring precise multi-step numerical reasoning, whereas tasks relying on conceptual or high-level understanding may saturate earlier.

Interestingly, for most of the open-source models, leveraging the chain-of-thought prompt hurt the overall performance; however, the Quantitative Reasoning ability improves obviously. For instance, the overall performance of Qwen2.5-VL-32B-Instruct drops 0.7\% but the Quantitative Reasoning increases from 11.43\% to 17.96\%, which is a 57.13\% relative boost. This demonstrates again the intricate reasoning cost in our quantitative questions, which requires multi-step calculation and can be obviously improved by the chain-of-thought prompt. 

\begin{figure}[h]
    \centering
    \includegraphics[width=1.0\textwidth]{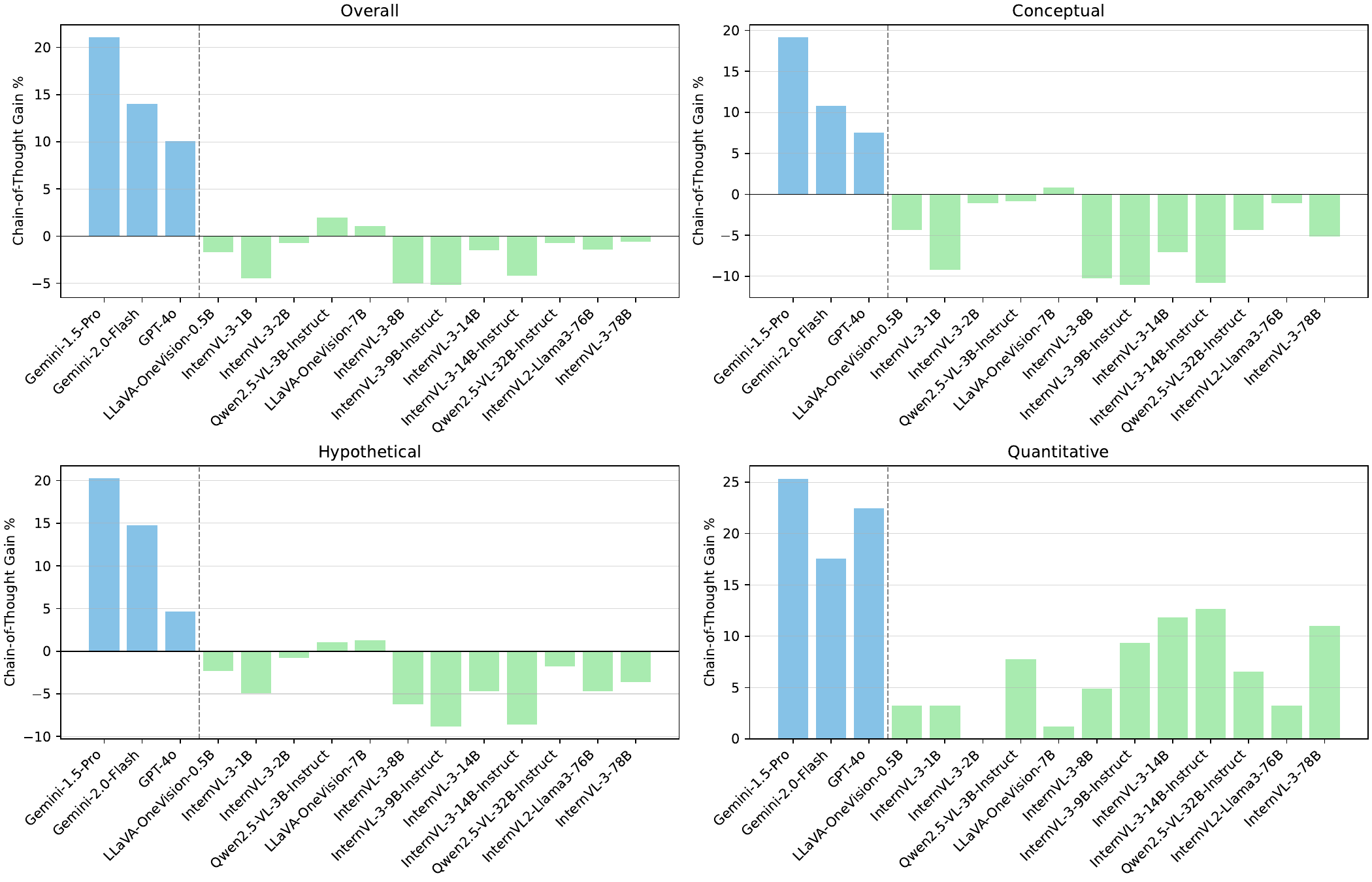} 
    \caption{Chain-of-Thought (CoT) performance gains across \textcolor{Proprietary}{proprietary models} and \textcolor{OpenSource}{open-source models}. 
    An obvious difference between \textcolor{Proprietary}{proprietary models} across all the reasoning aspects can be observed; 
    while for \textcolor{OpenSource}{open-source models}, quantitative reasoning has a notable performance boost, 
    while the other two reasoning aspects have negative impacts. 
    This phenomenon again demonstrates that the quantitative settings in ~\scbench require sophisticated multi-step reasoning 
    that can benefit a lot from chain-of-thought prompts. Best viewed in color.}
    \label{fig:cot_gain}
\end{figure}

To further investigate the possible reason that CoT prompting has a such different impact for different reasoning type for open-source models, we analyze the predictions of three representative models: InternVL-3-78B, InternVL2-Llama3-76B, and Qwen2.5-VL-32B-Instruct. From Table~\ref{tab:cot_vs_direct}, we observe that chain-of-thought (CoT) prompting does not consistently improve evaluation performance across reasoning types. Globally, the number of instances where the direct strategy is correct but CoT is wrong slightly exceeds the reverse, indicating that CoT often introduces additional errors. A more detailed breakdown by reasoning type reveals that the negative impact is concentrated in \textit{Conceptual} and \textit{Hypothetical} reasoning. In these categories, CoT tends to amplify hallucinations or over-elaborate on causal relations, leading the model away from the correct choice, whereas direct answering can rely more on memorized factual knowledge or surface-level pattern recognition. In contrast, \textit{Quantitative} reasoning benefits substantially from CoT: across the three models, we see more cases where CoT corrects errors that the direct approach would miss. This suggests that explicit reasoning chains help models externalize intermediate computational steps (e.g., arithmetic or logical comparisons), which are otherwise challenging when only a single final answer is required. A plausible reason why CoT hurts open-source models on conceptual and hypothetical reasoning is that open-source models are less robust in producing faithful long-form causal explanations, often hallucinating or over-elaborating when reasoning about abstract scenarios, but CoT provides a clear scaffold for step-by-step numerical comparisons, which directly reduces errors in quantitative tasks.

Compared to the CoT gains reported on other video reasoning benchmarks (using Gemini-2.0-Flash as an example) —MMVU~\cite{zhao2025mmvu} (+3.0\%), Video-Holmes~\cite{videoholmes} (+12.5\%), VideoMathQA~\cite{rasheed2025videomathqa} (+6.6\%), and MMR-V~\cite{zhu2025mmrv} (+2.4\%)—the +14.00\% improvement on \scbench is striking. This significant gap strongly indicates that \textbf{\scbench inherently demands more complex, multi-step reasoning than existing benchmarks, making it a more challenging and discriminative testbed for evaluating reasoning capabilities in multimodal models}.

\begin{tcolorbox}[takeaway,title=Takeaway: The Impact of Chain-of-Thought Prompt]
Chain-of-thought prompting consistently boosts quantitative reasoning for both proprietary and open-source models, while improvements in conceptual and hypothetical reasoning are observed almost only in proprietary models. 
\end{tcolorbox}

\begin{table*}[t]
\centering
\caption{Comparison of agreement and disagreement between CoT and Direct answers across three models.}
\resizebox{\textwidth}{!}{
\begin{tabular}{lccccc}
\toprule
\textbf{Model} & \textbf{Both Correct} & \textbf{Both Wrong Agree} & \textbf{Both Wrong Disagree} & \textbf{CoT Correct, Direct Wrong} & \textbf{Direct Correct, CoT Wrong} \\
\midrule
\multicolumn{6}{c}{\textbf{Overall}} \\
\midrule
InternVL-3-78B      & 259 & 253 & 242 & 120 & 126 \\
InternVL2-Llama3-76B      & 139 & 287 & 340 & 110 & 124 \\
Qwen2.5-VL-32B-Instruct      & 96  & 335 & 338 & 112 & 119 \\
\midrule
\multicolumn{6}{c}{\textbf{Conceptual Reasoning}} \\
\midrule
InternVL-3-78B      & 149 & 77  & 41  & 42  & 61  \\
InternVL2-Llama3-76B      & 58  & 125 & 97  & 43  & 47  \\
Qwen2.5-VL-32B-Instruct      & 37  & 147 & 92  & 39  & 55  \\
\midrule
\multicolumn{6}{c}{\textbf{Hypothetical Reasoning}} \\
\midrule
InternVL-3-78B      & 93  & 102 & 88  & 44  & 58  \\
InternVL2-Llama3-76B      & 48  & 114 & 129 & 38  & 56  \\
Qwen2.5-VL-32B-Instruct      & 49  & 145 & 106 & 39  & 46  \\
\midrule
\multicolumn{6}{c}{\textbf{Quantitative Reasoning}} \\
\midrule
InternVL-3-78B      & 17  & 74  & 113 & 34  & 7   \\
InternVL2-Llama3-76B      & 33  & 48  & 114 & 29  & 21  \\
Qwen2.5-VL-32B-Instruct      & 10  & 43  & 140 & 34  & 18  \\
\bottomrule
\label{tab:cot_vs_direct}
\end{tabular}}
\end{table*}


\subsection{The Impact of LMM Model Size}
We investigate the relationship between model size and performance across four representative open-source model families: InternVL-3, InternVL-2, and Qwen2.5-VL.
In general, a positive correlation between parameter count and overall performance can be observed within each series; however, larger size does not universally guarantee higher performance, and diminishing or even negative returns occur in some cases.

For InternVL-3, the trend from smaller to larger configurations shows substantial gains in overall accuracy (14.0\% $\rightarrow$ 35.7\%). The largest jump occurs when scaling from 1B to 7B, where overall performance increases by over 11 percentage points, with notable improvements in Conceptual (16.76\% $\rightarrow$ 43.78\%) and Medicine (14.02\% $\rightarrow$ 34.58\%). However, gains are not strictly monotonic. For instance, the 9B variant (27.20\%) underperforms compared to the 8B-Instruct variant (29.40\%) in Overall, despite having more parameters, partly because of different language backbones.
For InternVL-2, performance improvements with scale are modest and less consistent. The overall accuracy rises from 14.4\% (0.5B) to 21.3\% (4B) but then plateaus, with the 20B model achieving only 19.5\%. The 4B model outperforms the 7B variant in all question types except Quantitative, suggesting that architectural or training differences outweigh the size advantage.
For Qwen2.5-VL, the scaling law appears weaker. The 3B, 7B, and 32B models show modest overall performance increases (16.1\%, 16.4\%, 21.5\%), but the 72B variant drops slightly to 20.3\%. This plateauing, and in some cases regression, is particularly evident in Conceptual and Chemistry performance, where the 72B model lags behind the 32B variant. However, the 72B model shows strong Quantitative Reasoning, which requires advanced multi-step reasoning.

However, between different model series, a larger model does not always guarantee better performance. For instance, Qwen2.5-VL-72B-Instruct is not even as good as InternVL2-4B in overall performance.
Overall, while model size generally correlates with improved performance, especially in Conceptual and domain-specific disciplines (e.g., Medicine, Biology), the relationship is far from linear or absolute. Variations between close sizes—such as InternVL-3 7B vs. 8B, or Qwen2.5-VL 32B vs. 72B—\textbf{underscore that scaling must be accompanied by corresponding improvements in data quality, training objectives, and architectural choices to realize consistent gains in our \scbench}.

\begin{figure}[H]
    \centering
    \includegraphics[width=1.0\textwidth]{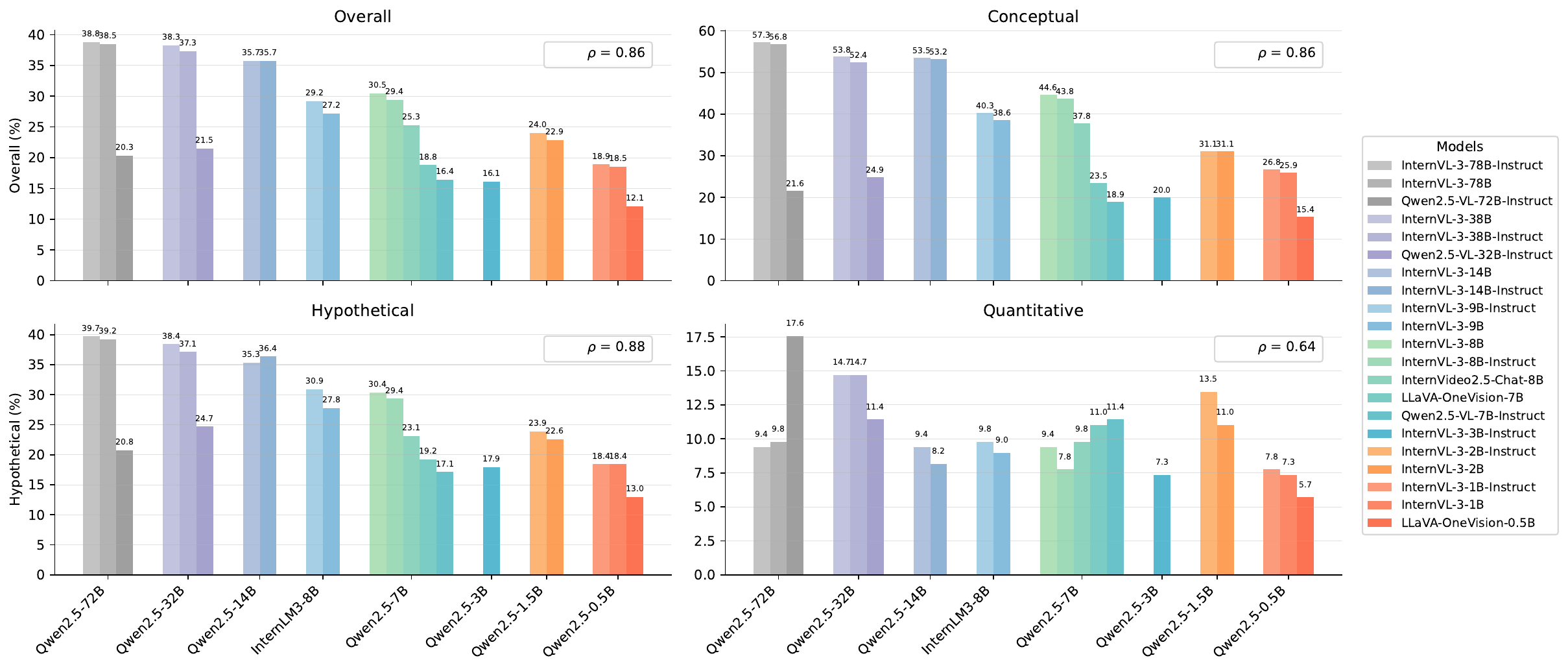} 
    \caption{The impact of LMM backbones on the performance. }
    \label{fig:llm_size}
\end{figure}

\subsection{The Impact of LLM Backbone Size}
We comprehensively compare the performance across LMMs with different LLM backbones as shown in Figure~\ref{fig:llm_size}. In general, we can observe a clear positive correlation between the size of the LLM backbone and the overall performance of the corresponding LMMs. Larger backbones such as Qwen2.5-72B and Qwen2.5-32B consistently achieve the highest scores, while smaller backbones (e.g., Qwen2.5-0.5B, Qwen2.5-1.5B) exhibit significantly weaker performance. This trend highlights the critical role of language backbone capacity in driving the multimodal reasoning ability of LMMs.

Interestingly, the correlation is strong in the \textit{conceptual} and \textit{hypothetical} reasoning ($\rho = 0.86$ and $\rho = 0.88$, respectively), where scaling the backbone size almost monotonically improves performance for most of the VLM series (e.g., InternVL-3 series).
In contrast, the \textit{quantitative} reasoning exhibits a much weaker correlation ($\rho = 0.64$): simply increasing the backbone size does not yield proportional gains. 
This indicates that the difficulty of quantitative reasoning cannot be resolved by scaling language ability alone, but likely requires more accurate visual perception and advanced numerical reasoning capability.

\begin{tcolorbox}[takeaway,title=Takeaway: The Impact of Model Scaling]
Larger models reliably boost conceptual/hypothetical reasoning, but quantitative gains remain weak and non-monotonic across series.
\end{tcolorbox}

\subsection{The Impact of Audio}
To investigate how audio, which provides additional information about the experiments shown in the video, will affect the model performance, we evaluate Gemini-2.5-Pro~\cite{google2025gemini25pro} and Qwen2.5-Omni-7B~\cite{Qwen2.5-Omni} due to their capability of supporting full modality input. From Table~\ref{tab:audio}, we can observe consistent improvement brought by audio input for both Gemini-2.5-Pro and Qwen2.5-Omni-7B in overall performance (2.70\% and 2.80\%, respectively). This limited improvement also demonstrates that the visual content dominates the model performance.

\begin{table*}[htbp]
\centering
\caption{Model Performance Improvement by Audio on \scbench.}
\resizebox{\textwidth}{!}{%
\begin{tabular}{@{}lccccccccccc@{}}
\toprule
\multirow{2}{*}{\textbf{Models}} & 
\multirow{2}{*}{\textbf{Use Audio}} & 
\multirow{2}{*}{\textbf{Overall}} & 
\multicolumn{3}{c}{\textbf{Question Type}} & 
\multicolumn{4}{c}{\textbf{Discipline}} \\
\cmidrule(lr){4-6} \cmidrule(lr){7-10}
& & & \textbf{Conceptual} & \textbf{Hypothetical} & \textbf{Quantitative} &
\textbf{Biology} & \textbf{Chemistry} & \textbf{Medicine} & \textbf{Physics} \\
\midrule
Gemini-2.5-Pro~\cite{google2025gemini25pro} & N & 64.30 & 69.73 & 67.79 & 50.61 & 64.79 & 61.82 & 74.77 & 61.44 \\
Gemini-2.5-Pro~\cite{google2025gemini25pro} & Y & 67.00 & 74.32 & 67.53 & 55.10 & 68.70 & 61.21 & 71.96 & 66.14 \\
Qwen2.5-Omni-7B~\cite{Qwen2.5-Omni} & N & 14.70 & 16.69 & 14.97 & 9.74 & 14.39 & 14.36 & 17.85 & 14.25 \\
Qwen2.5-Omni-7B~\cite{Qwen2.5-Omni} & Y & 17.50 & 20.29 & 18.51 & 12.80 & 18.24 & 16.52 & 18.19 & 17.67 \\
\bottomrule
\end{tabular}
}
\label{tab:audio}
\end{table*}

\subsection{Error Analysis}
To gain an in-depth understanding of how far the current models are from real human experts, we comprehensively analyze the results of Gemini-1.5-Pro-Thinking and Gemini-2.0-Flash-Thinking. Comparing these reasoning steps with the real rationale from the annotation process, which was carefully verified by human experts, we found three noticeable errors: Incorrect Visual Perception, Inaccurate Reasoning, and Lack of Domain Knowledge. Importantly, most of the wrong responses stem from a complex error combination, e.g., both incorrect visual perception and inaccurate reasoning progress.

\noindent\textbf{Incorrect Visual Perception (70.68\%)} is the most common factor that leads to an incorrect conclusion. It leads to misinterpret of what is visually shown in the video and identify the wrong moment in the temporal span. For instance, in Figure~\ref{fig:failure}, Gemini-2.0-Flash overlooks the on-screen textual information indicating the humidity, which leads to the consequence that it obtains incorrect option analysis for option J, maintaining standardized high humidity.

\noindent\textbf{Inaccurate Reasoning Progress (63.25\%)} also constitutes a main portion of the error cause.  It usually involves logical flaws despite correct or partially correct observations. In Figure~\ref{fig:failure}, Gemini-2.0-Flash fails to establish the connection between the operation of using potassium sulfate and the high-humidity consequence. 

\noindent\textbf{Lack of Domain Knowledge (49.40\%)} happens when models fail to connect the visual evidence with specific expert knowledge that helps answer the question. For example, in Figure~\ref{fig:failure}, the function of potassium sulfate should be a well-known expert knowledge for researchers; however, Gemini-2.0-Flash is not able to analyze even perceive the existence of the sulfate in the video, which is strong evidence of domain knowledge lacking.

\subsection{Failure Case Analysis}
From Table~\ref{tab:main}, we observe that even the best-performing proprietary model, Gemini-2.5-Pro, achieves limited accuracy on \scbench, revealing the capped reasoning capabilities of current MLLMs in complex scientific experimental scenarios. Chain-of-thought (CoT) evaluation provides clearer insight into these reasoning limitations when facing highly specialized, visually grounded questions.

\begin{figure}[htbp]
\centering
\vspace{-0.2cm}
\scalebox{1.}{\includegraphics[width=0.8\textwidth]{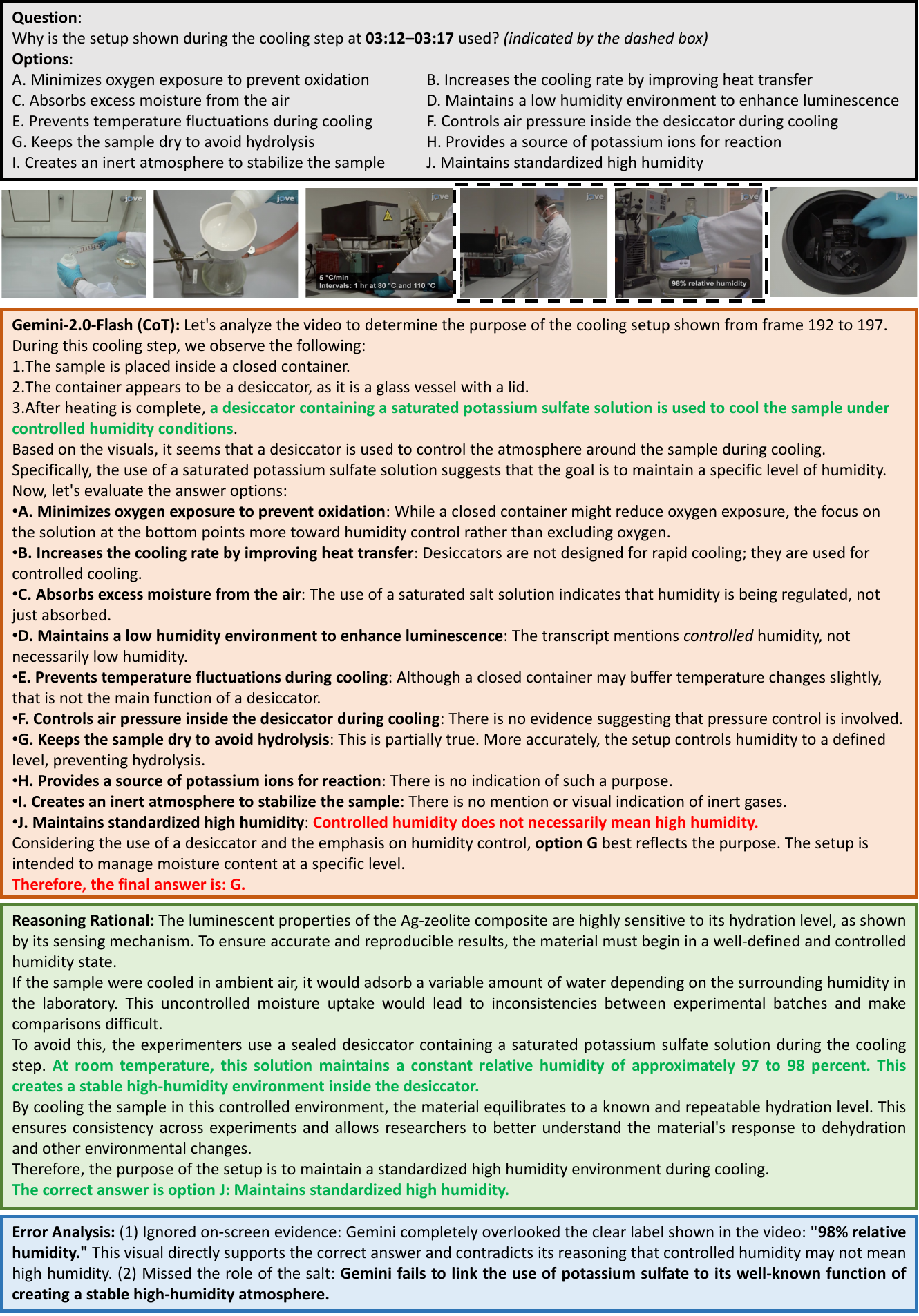}}
\caption{Failure case example of Gemini-2.0-Flash-Thinking.}
\label{fig:failure}
\vspace{-10pt}
\end{figure}

For example, in the case shown in Figure~\ref{fig:failure}, the question asks: “Why is the setup shown during the cooling step at 03:12–03:17 used?” The correct rationale is as follows: the setup is a sealed desiccator containing a saturated potassium sulfate solution, which maintains a constant relative humidity of approximately 98\%. This high-humidity environment is crucial because the luminescent properties of the Ag-zeolite composite are highly sensitive to hydration. Cooling the sample in this controlled atmosphere ensures it equilibrates to a well-defined hydration state, avoiding batch-to-batch variation caused by fluctuating laboratory humidity. The correct answer is J (Maintains standardized high humidity).
However, Gemini-2.0-Flash (CoT) fails despite partially recognizing that the desiccator controls humidity. It incorrectly rules out “standardized high humidity” and instead selects G (Keeps the sample dry to avoid hydrolysis). This mistake stems from two major reasoning errors: (1) Ignored on-screen evidence – The model completely overlooked the clear text overlay “98\% relative humidity” in the video, which directly confirms the correct answer. (2) Misinterpreted the chemical role of potassium sulfate – The model failed to associate potassium sulfate’s well-established function with generating a stable high-humidity environment, instead generalizing it as generic moisture control.
This example illustrates that solving such questions requires a combination of precise temporal localization (identifying the cooling step), fine-grained spatial perception (reading on-screen labels), and expert-level domain knowledge (understanding the chemical’s function in environmental control). Current models, even with CoT prompting, often break down when any one of these components is missing.



\section{Related Work}
\subsection{Video Large Language Models}
Recent advances in large multimodal models have led to the rapid development of LMMs that empower temporal perception and reasoning ability. VideoChatGPT~\cite{maaz2023videochatgpt} leverages a simple spatiotemporal pooling mechanism for the visual encoding, while the LLaVA~\cite{liu2023visual,li2024llavanext} series leverage \textit{anyres} technique to unify both image and video encoding. Meanwhile, LongVA~\cite{zhang2024longva} and LongVILA~\cite{chen2024longvila} aim at processing long-term video by extending the context window of LLMs in a multi-stage training pipeline. However, their ability to perform scientific reasoning, such as interpreting experimental procedures, analyzing quantitative measurements, remains underexplored. To address this gap, we propose \scbench, a domain-specific benchmark designed to evaluate video reasoning in real-world scientific scenarios.

\subsection{Video Reasoning Benchmarks}
Recent benchmarks have increasingly emphasized complex temporal reasoning and multimodal understanding to evaluate LMMs beyond the traditional video QA benchmark paradigm~\cite{yu2019activitynetqa,mangalam2023egoschema,MSRVTT-QA-MSVD-QA,xiao2021next,wu2024star}. Perception Test~\cite{patraucean2023perception} examines multiple reasoning modes such as descriptive, predictive, and counterfactual reasoning in real-world scenarios. MVBench~\cite{li2024mvbench} proposes 20 temporally grounded challenges that require understanding actions, object interactions, and motion cues across multiple frames. Video-MME~\cite{videomme} incorporates human annotations across videos ranging from 11 seconds to 1 hour, including subtitle and audio streams to promote multi-source comprehension. MMBench-Video targets long-context temporal reasoning by leveraging hour-long videos with open-ended QA. Domain-specific benchmarks have also emerged. WorldQA~\cite{zhang2024worldqa} focuses on long-chain reasoning with multimodal world knowledge, Video-MMMU~\cite{videommmu} collects expert-level instructional videos across disciplines for multi-stage knowledge acquisition, and Video-MMLU~\cite{song2025videommlu} centers on lecture-level understanding in math, physics, and chemistry. Our work extends this line of research by introducing a scientific video reasoning benchmark grounded in real experimental settings, emphasizing measurement, calculation, and conceptual reasoning involved in specific experiments.


\subsection{AI for Science}

In recent years, artificial intelligence has made tremendous progress across a variety of scientific disciplines, including biology~\cite{alphafold,yang2023alphafold2,esmfold}, chemistry~\cite{qiao2020orbnet,bran2023chemcrow}, materials science~\cite{kim2023gnome,chen2023graph}, and mathematics~\cite{lewkowycz2022solving}. These breakthroughs have rapidly pushed the boundaries of scientific research and demonstrated performance that rivals, and in some cases surpasses, traditional methods.
For instance, AlphaFold~\cite{alphafold,yang2023alphafold2} revolutionized protein structure prediction by achieving near-experimental accuracy through deep learning, enabling the release of structures for nearly all known proteins. In materials science, GNoME~\cite{kim2023gnome} employed graph neural networks to discover over 700{,}000 stable crystal structures, dramatically expanding the known materials space. In chemistry, models like OrbNet~\cite{qiao2020orbnet} and ChemCrow~\cite{bran2023chemcrow} integrate quantum chemistry and symbolic tools with language models to enable efficient simulation and synthesis planning. 
Despite these advances, current AI systems remain far from achieving the capabilities of a versatile researcher who can perform complex scientific tasks across domains with expert-level depth and accuracy. In this work, we introduce \scbench, a challenging benchmark that focuses on scientific video reasoning, requiring visual perception, domain-specific knowledge, and intricate reasoning. Our benchmark aims to assess how far current LMMs are from expert-level scientific reasoning and to foster the development of next-generation AI systems for science.

\section{Conclusion}

We present \scbench, the first scientific video reasoning benchmark that demands PhD-level knowledge to perform complex reasoning grounded in real-world experimental scenarios. \scbench bridges the gap between general-purpose video understanding benchmarks and advanced scientific reasoning. It spans four major scientific disciplines—\textit{Physics}, \textit{Chemistry}, \textit{Biology}, and \textit{Medicine}—and covers over 20 distinct subjects, encompassing a broad spectrum of scientific inquiry.
To enable rigorous evaluation, we design three complementary question types—quantitative, hypothetical, and conceptual—that reflect common reasoning challenges in scientific research. We evaluate a total of 21 models, including 6 proprietary and 15 open-source LMMs, to assess their current capabilities relative to expert human reasoning. Our results show that even the most advanced model (Gemini-2.5-Pro) still struggles with these challenging tasks.
To further explore the impact of reasoning augmentation, we evaluate model performance under chain-of-thought prompting. The results demonstrate clear performance gains, highlighting the importance of explicit reasoning in complex scientific domains.
We hope \scbench will serve as a milestone benchmark for advancing LMMs and inspire future research in the AI for Science community.

{
    \small
    \bibliography{main}
}

\clearpage
\setcounter{page}{1}
\appendix

\section{Appendix overview}
This document provides more details of our approach and additional experimental results, organized as follows:
\begin{itemize}
	\vspace{2pt}
 	\item \S~\ref{discipline} Discilpines and Subjects in \scbench
        \item \S~\ref{config} Detailed configuration of evaluated models.
        \item \S~\ref{metadata} Gemini-assisted Metadata Generation
 	\item \S~\ref{more_failure}  More Failure Cases.

\end{itemize}

\section{Disciplines and Subjects}
\label{discipline}
\begin{table*}[htbp]
\centering
\caption{List of subjects grouped by their corresponding disciplines.}
\begin{tabular}{ll}
\toprule
\textbf{Discipline} & \textbf{Subjects} \\
\midrule
Biology &
Biochemistry, Bioengineering, Biogeotechnology, Bioinformatics,  \\
& Genetic Engineering, Lipidomics, Mechanobiology, Microfluidics,  \\
& Mycology, Neurohistology, Neuroscience, Phycology, Proteomics, \\
& Regenerative Biology, Structural Biology, Cell Biology, Molecular Biology  \\
\midrule
Chemistry &
Analytical Chemistry, Electrochemistry, Green Chemistry, \\
& Nanomaterials, Organic Chemistry, Photocatalysis, Photovoltaics, \\
& Physical Chemistry, Radiochemistry, Materials Chemistry \\
\midrule
Medicine &
Biomaterials, Cardiovascular Research, Dentistry, Drug Delivery, Immunology, \\
& Molecular Imaging, Nanomedicine, Oncology, Ophthalmology, Pharmacology, \\
& Radiopharmaceuticals, Regenerative Medicine, Surgical Devices, Therapeutics \\
\midrule
Physics &
Acoustofluidics, Additive Manufacturing, Aerosol Science, Applied Physics, \\
& Ceramic Engineering, Ceramics, Civil Engineering, Condensed Matter Physics, \\
& Ecological Engineering, Electronics, Fluid Dynamics, Materials Science, \\
& Mechanical Engineering, Microfluidics, Nanofabrication, Plasma Physics, \\
& Semiconductor, Sensors, Soft Robotics, Thermal Engineering \\
\bottomrule
\end{tabular}
\end{table*}

\clearpage
\section{Configuration of Evaluated Models}
\label{config}

All of the evaluated models in this paper were released after May 2024. The detailed configuration of the evaluated models are shown in Table~\ref{model_config}. We follow the default frame sampling settings of the official implementation of each model. The temperature is set to 0 to ensure stable predictions.

\begin{table*}[htbp]
\centering
\caption{Models used in our evaluation with confirmed release dates where available.}
\renewcommand{\arraystretch}{1.1}
\setlength{\tabcolsep}{6pt}
\begin{tabular}{l l l l}
\toprule
\textbf{Organization} & \textbf{Model} & \textbf{Release Date} & \textbf{Input Frames}\\
\midrule
\multicolumn{4}{l}{\textit{Proprietary Models}} \\
\midrule
Google   & Gemini-2.5-Pro              & June 17, 2025        &  1fps \\
Google   & Gemini-2.5-Flash            & June 17, 2025        &  1fps \\
Google   & Gemini-2.0-Flash            & February 5, 2025     &  1fps \\
Google   & Gemini-1.5-Pro              & September 24, 2024   &  1fps \\
OpenAI   & GPT-4o                      & May 13, 2024         &  256 \\
\midrule
\multicolumn{4}{l}{\textit{Open-source Multimodal Models}} \\
\midrule
OpenGVLab  & InternVL-3 series     & April 11, 2025       &  32 \\
Alibaba  & Qwen2.5-VL     & January 28, 2025     &  768 \\
OpenGVLab & InternVideo2.5               & January 21, 2025     & 512 \\
LMMS-Lab & LLaVA-OneVision              & August, 2024         &  32 \\
OpenGVLab  & InternVL-2 series     & July 4, 2024         & 16 \\
LMMS-Lab & LongVA (7B)                   & June 24, 2024    & 128 \\
UCSD/CMU & LLaVA-NeXT-Video          & May 10, 2024         &  32\\
\bottomrule
\label{model_config}
\end{tabular}
\end{table*}

\clearpage
\section{Metadata Generation Using Gemini 2.5 Pro}
\label{metadata}

\begin{figure}[htbp]
\centering
\vspace{-0.2cm}
\scalebox{1.}{\includegraphics[width=0.85\textwidth]{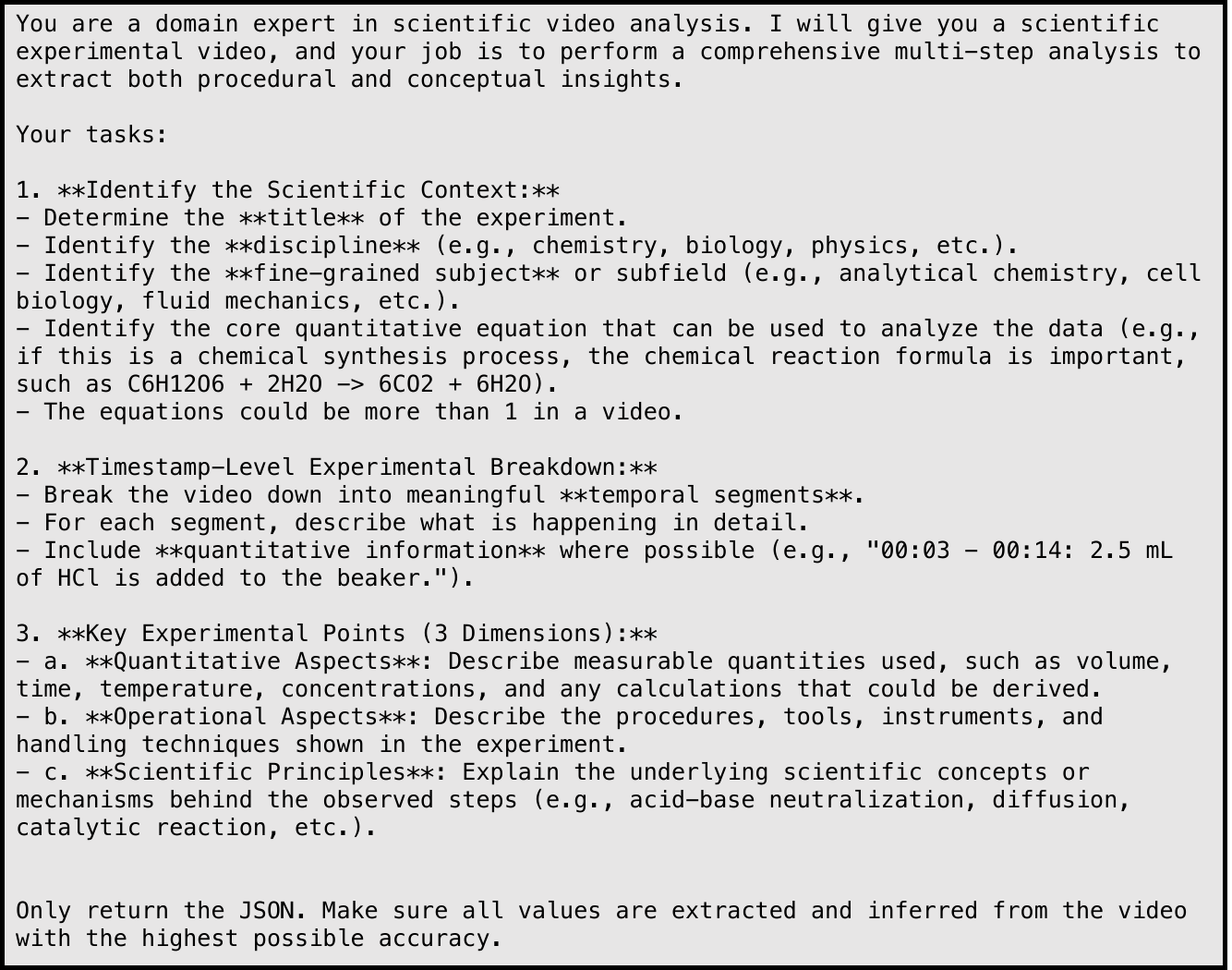}}
\caption{The prompt for Gemini 2.5 Pro to generate metadata to support initial annotation.}
\label{fig:meta_prompt}
\vspace{-10pt}
\end{figure}

\begin{figure}[htbp]
\centering
\vspace{-0.2cm}
\scalebox{1.}{\includegraphics[width=0.85\textwidth]{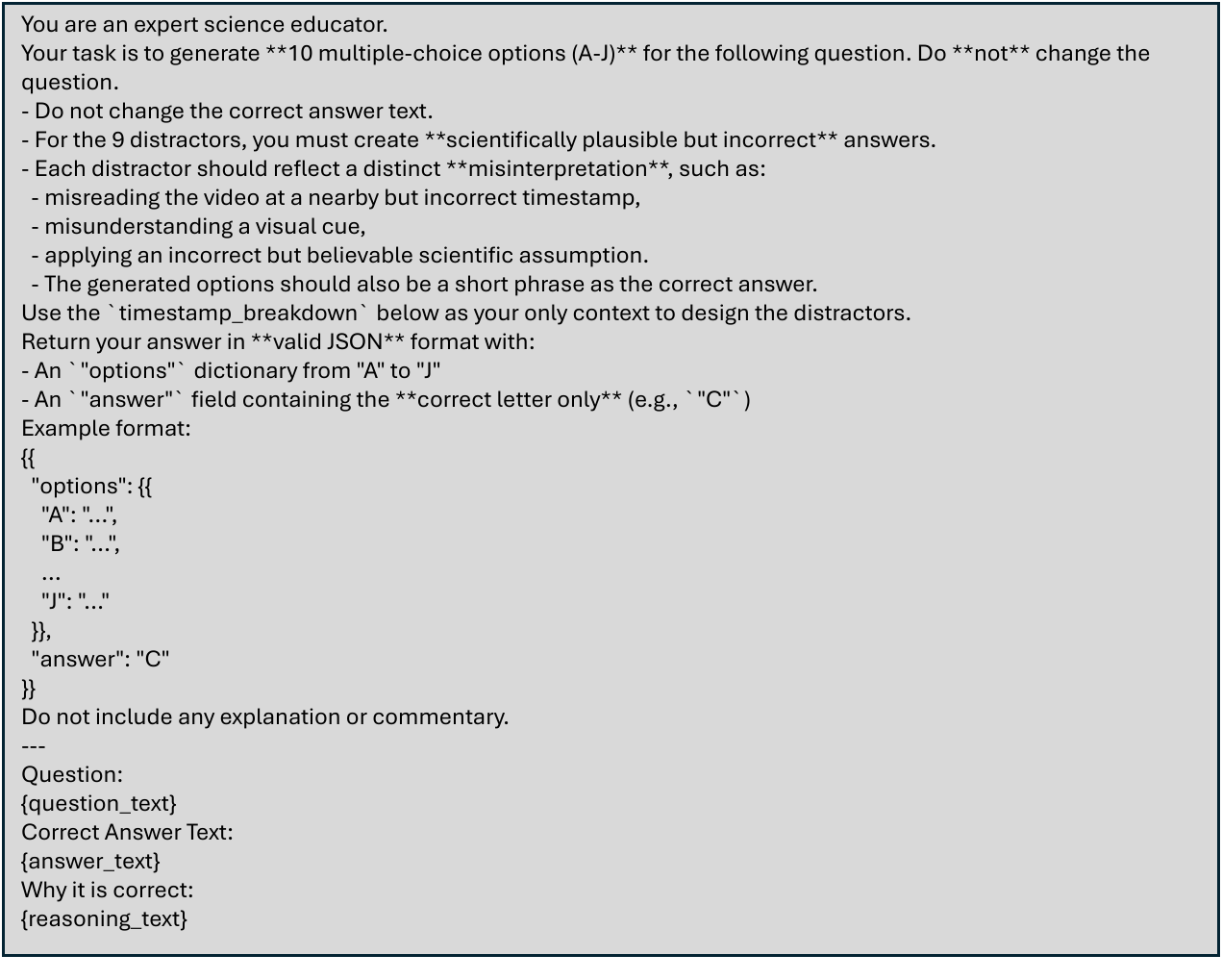}}
\caption{The prompt for Gemini 2.5 Pro to generate distractors.}
\label{fig:opt_gen}
\vspace{-10pt}
\end{figure}

\begin{figure}[htbp]
\centering
\vspace{-0.2cm}
\scalebox{1.}{\includegraphics[width=0.85\textwidth]{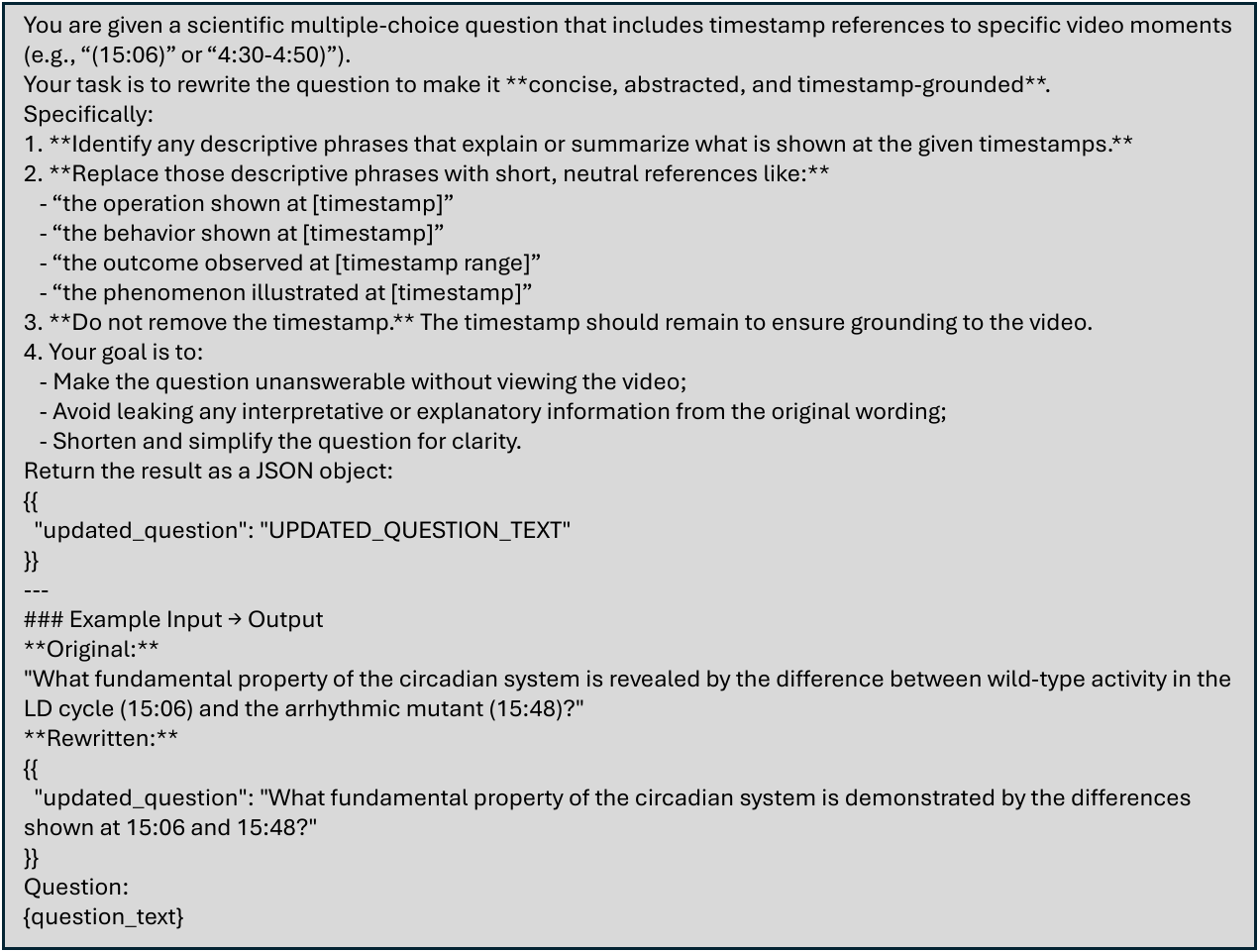}}
\caption{The prompt for Gemini 2.5 Pro to update distractors.}
\label{fig:opt_update}
\vspace{-10pt}
\end{figure}

\clearpage

\lstset{
  basicstyle=\ttfamily\footnotesize,
  breaklines=true,
  backgroundcolor=\color{gray!10},
  frame=single
}

\lstdefinelanguage{json}{
  basicstyle=\ttfamily,
  numbers=left,
  numberstyle=\tiny,
  stepnumber=1,
  numbersep=5pt,
  showstringspaces=false,
  breaklines=true,
  frame=lines,
  backgroundcolor=\color{gray!10},
  string=[s]"",
  morestring=[b]',
  literate={:}{{:}}{1},
}

\begin{lstlisting}[language=json, caption={Example metadata generated from Gemini 2.5 Pro}]
{
  "66497": {
    "title": "Synthesis and Characterization of Self-Assembled Metal-Organic Framework Monolayers Using Polymer-Coated Particles",
    "discipline": "Chemistry",
    "subject": "Materials Chemistry",
    "core_equation": "UiO-66-DDMAT + n(CH2=CHCOOCH3) --(Photocatalyst, Light)--> UiO-66-p(CH2-CH(COOCH3))n",
    "timestamp_breakdown": [
      "00:00 - 01:58: An introduction to the research...",
      "01:59 - 02:05: 10 mg of catechol-DDMAT is weighed...",
      "02:05 - 02:10: 5 mL of chloroform is added...",
      "...",
      "06:13 - 06:48: After the toluene evaporates, a freestanding monolayer forms..."
    ]
  }
}
\end{lstlisting}

\clearpage
\section{More Failure Case Studies}
\label{more_failure}

In this section, we showcase more failure case among Conceptual, Hypothetical, and Quantitative Reasoning.

\paragraph{Case 1: Misinterpretation of Spectral Evidence.} 
InternVL-3-14B misread the electroluminescence spectrum by hallucinating multiple peaks and voltage-dependent shifts, concluding that multiple emissive species contributed to the emission. In reality, the spectrum displayed a single stable peak, indicating well-confined recombination. The model failed to recognize confinement as the key property and instead introduced instability, leading to the wrong answer. This reflects a tendency to over-generalize from prior knowledge of spectroscopy rather than grounding reasoning in the observed evidence.

\paragraph{Case 2: Misunderstanding Experimental Procedure.}
Qwen2.5-VL-32B erred in interpreting the role of a 20-minute incubation step. It concluded that failure of incubation would prevent substrate hydrolysis, even though the substrate was not yet present during this period. The correct rationale emphasized pre-incubation of inhibitors with elastase, ensuring binding equilibrium before substrate addition. The model ignored this inhibitor-binding context, conflating incubation with general reaction progress, which resulted in selecting an irrelevant option.

\paragraph{Case 3: Incorrect Solvent Accounting.}
InternVL-3-9B miscalculated the total solvent volume in a polymerization setup. It fabricated multiple additions of 1,4-dioxane and neglected the actual solvents present in the protocol (DMSO and DMF). Furthermore, it mishandled the inclusion of solvent-containing stock solutions and confused monomer additions with solvents. This misidentification led to both an incorrect summation and a mismatch between its numerical reasoning and the final selected option. The correct calculation, grounded in protocol details, yielded a total of 4.462 mL, which the model overlooked.

\paragraph{Conclusion.}
Across these cases, a consistent pattern emerges: models often fail due to misalignment between the experimental context and the reasoning they generate. Errors stem from three main sources: (1) hallucination of experimental features not present in the input (spectral peaks, solvent additions), (2) failure to account for the temporal sequence of steps (substrate not present during incubation), and (3) confusion between categories of experimental components (monomers vs. solvents). These findings highlight the importance of grounding reasoning strictly in the provided experimental evidence rather than relying on generic domain priors. Improved training on step-by-step alignment between protocol details and answer derivation may reduce these systematic failures.

\begin{figure}[htbp]
\centering
\vspace{-0.2cm}
\scalebox{1.}{\includegraphics[width=0.95\textwidth]{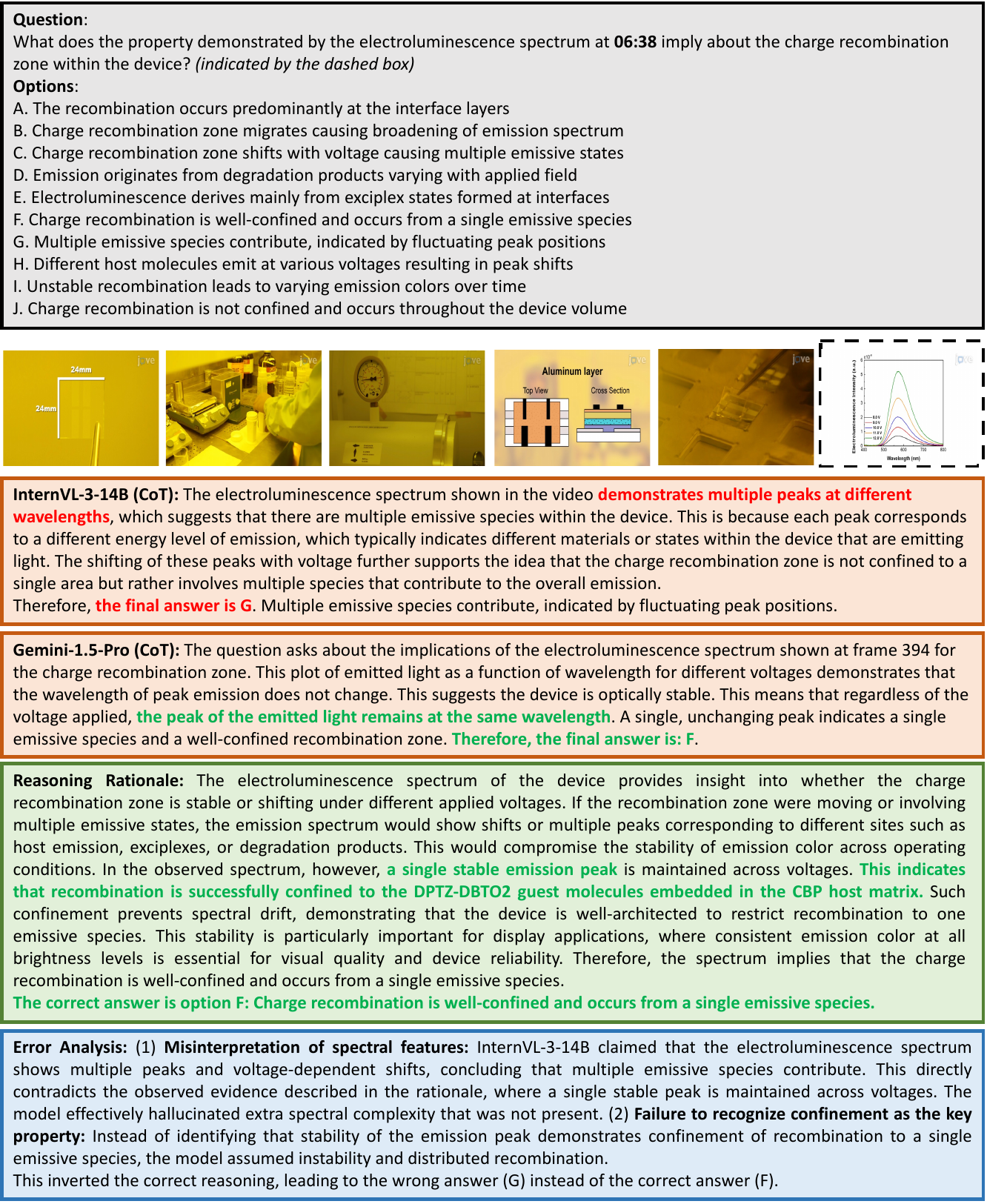}}
\caption{Failure case example of InternVL-3-14B with chain-of-thought prompt compared with the correct reasoning of Gemini 1.5 Pro. InternVL-3-14B makes the wrong predictions because of the incorrect visual perception of the experiment results, while Gemini 1.5 Pro correctly captures the unchanged state of the peak at different wavelength.}
\label{fig:vl3_14b_failure}
\vspace{-10pt}
\end{figure}

\begin{figure}[htbp]
\centering
\vspace{-0.2cm}
\scalebox{1.}{\includegraphics[width=0.75\textwidth]{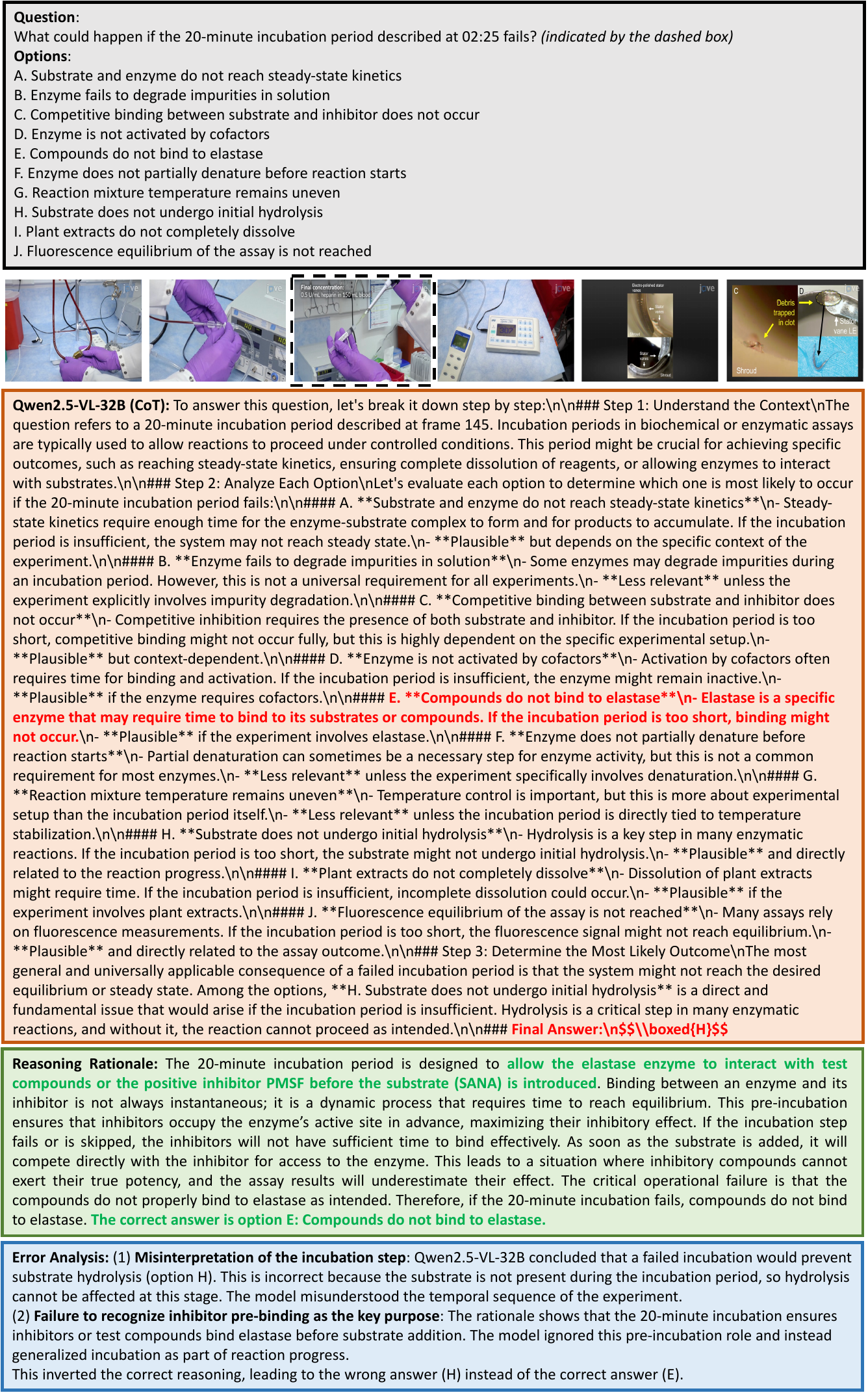}}
\caption{Failure case example of Qwen2.5-VL-32B-Instruct. The model makes the wrong prediction because of lacking the domain knowledge about the incubation process. }
\label{fig:failure}
\vspace{-10pt}
\end{figure}

\begin{figure}[htbp]
\centering
\vspace{-0.2cm}
\scalebox{1.}{\includegraphics[width=0.85\textwidth]{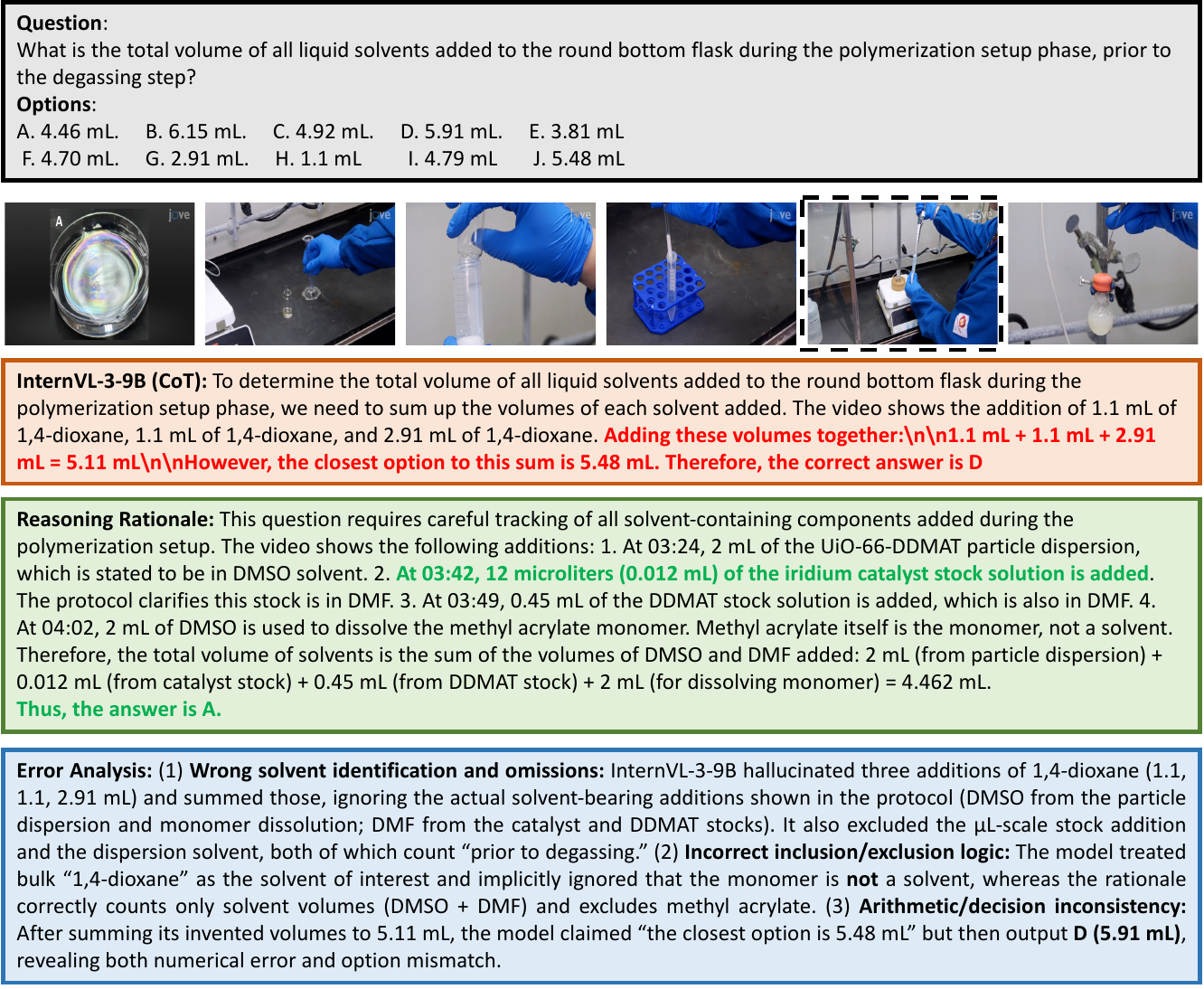}}
\caption{Failure case example of InternVL-3-9B. The model makes the wrong predictions because of the wrong calculating logic and wrong solvent identification.}
\label{fig:failure}
\vspace{-10pt}
\end{figure}

\end{document}